# TextPixs: Glyph-Conditioned Diffusion with Character-Aware Attention and OCR-Guided Supervision

**Syeda Anshrah Gillani**[*]
Hamdard University
syedaanshrah16@gmail.com

**Mirza Samad Ahmed Baig**[†]
Danat Fz LLC (Argaam)
mirzasamadcontact@gmail.com

**Osama Ahmed Khan**[*]
Hamdard University
osama.ahmed@hamdard.edu.pk

**Shahid Munir Shah**[*]
Hamdard University
shahid.munir@hamdard.edu.pk

**Umema Mujeeb**[*]
Hamdard University
umemamujeeb905@gmail.com

**Maheen Ali**[*]
Hamdard University
maheenasif494@gmail.com

[*]Hamdard University, Karachi, Pakistan
[†]Danat Fz LLC (Argaam), Karachi, Pakistan

## Abstract

The modern text-to-image diffusion models boom has opened a new era in digital content production as it has proven the previously unseen ability to produce photorealistic and stylistically diverse imagery based on the semantics of natural-language descriptions. However, the consistent disadvantage of these models is that they cannot generate readable, meaningful, and correctly spelled text in generated images, which significantly limits the use of practical purposes like advertising, learning, and creative design. This paper introduces a new framework, namely **Glyph-Conditioned Diffusion with Character-Aware Attention (GCDA)**, using which a typical diffusion backbone is extended by three well-designed modules. To begin with, the model has a **dual-stream text encoder** that encodes both semantic contextual information and explicit glyph representations, resulting in a **character-aware representation** of the input text that is rich in nature. Second, an attention mechanism that is aware of the character is proposed with a new attention segregation loss that aims to limit the attention distribution of each character independently in order to avoid distortion artifacts. Lastly, GCDA has an **OCR-in-the-loop fine-tuning** phase, where a full text perceptual loss, directly optimises models to be legible and accurately spell. Large scale experiments to benchmark datasets, such as MARIO-10M and T2I-CompBench, reveal that GCDA sets a new state-of-the-art on all metrics, with better character based metrics on text rendering (Character Error Rate: 0.08 vs 0.21 for the previous best; Word Error Rate: 0.15 vs 0.25), human perception, and comparable image synthesis quality on high-fidelity (FID: 14.3).

---

[*]Corresponding author: Syeda Anshrah Gillani (syedaanshrah16@gmail.com)

# 1 Introduction

Capabilities of modern text-to-image diffusion models are impressive, but one of the most popular and long-standing limitations remains a barrier to their practical use: the inability to produce coherent and correctly spelled text in generated images. Such models are capable of generating extremely photorealistic imagery, with a high level of fine-grained detail, including lighting, texture and scene composition. However, when given the opportunity to produce textual messages, they will usually come out distorted, unreadable, or meaningless (in the case of rendering a simple word like PIZZA on a sign, HELLO on a storefront). This mismatch between visual faithfulness and textual accuracy is well acknowledged in the academic literature since the advent of early diffusion-based generative models. It has often been recognized and debated, but it has mostly remained unresolved in several generations of progressively more advanced systems, such as DALL·E 2, Midjourney and Stable Diffusion. These models are very good at semantic scene synthesis, and able to create fantastical images given complex prompts, like "a steampunk octopus playing chess with Einstein in a library full of floating books", but always fail miserably at reproducing textual features. The present paper fills this gap directly by suggesting a specific architectural solution to increase the fidelity of text generation at the cost of the overall image quality.

The given weakness is especially exasperating considering the otherwise good language comprehension that these models have. It is not a matter of semantic understanding deficiency, in fact, such systems are quite efficient at understanding and answering complex textual stimuli. Instead, the issue is more complex and, as our analysis indicates, it lies in certain inherent architectural and representational constraints that have been mostly ignored in the previous studies.

We assume that language models have already reached the significant semantic competence; in other words, they can skillfully comprehend the meaning of the words both in the isolation and in the context. However, such systems are unable to perceive lexical forms as visual objects. Take an example of CLIP, the text encoder that is generally used to power such applications: when the system is run on the word OPEN, it will rightly find semantic vectors that refer to the concepts accessible, entrance, and available. More importantly, however, it is unaware that these notions are visualized in a particular lexical structure, i.e., the sequence O-P-E-N.

The principle of BPE-based segmentation is one of the fundamental mechanisms of transformer language models. BPE allows these models to model semantic patterns much more accurately than character-based approaches, by breaking input sequences into smaller units. Nonetheless, visual perception of words is seriously impaired by the same segmentation strategy. The fact that the same tokenisation of OPEN into OP and EN or even more randomly chosen pairs, such as effectively destroys any locally consistent sequences of letters.

We started to investigate the issue about two years ago, and it did not take us long to realize the issue was complex. The issue is relevant to underlying processes of present-day models of image-generation, such as how text is encoded, how images are produced, and how the models are trained.

A number of strategies have been tested but none has been completely satisfactory. These previous strategies are explained in detail in our companion publication dealing with design and evaluation. Our final decision is presented in the current work and it was formed in the result of the critical literature review and empirical investigation. Rather than face a head-on attack of the extant models, which can be successfully applied only under the condition that semantic knowledge is provided, we chose to consider both semantic and visual attributes within one framework.

We designed an approach that we name GCDA which implements three major components addressing various elements of the issue:

To begin with, we created a two-stream text encoder. Semantic processing is performed by one stream in the same way that exists with the current models but another stream interprets the actual visual appearance of characters. This makes sure that the model does not forget how the letters are like.

Second, we placed a spatial separation between successive individual characters in generation by using an attention mechanism. This will avoid the usual fate of the letters becoming a messy illegible blob.

Third, we implemented an OCR based feedback loop that directly grades the model on text accuracy and assists the model with improving. It is like having a spelling teacher going over the shoulder of the model.

The results obtained empirically are significantly better than those predicted by us. The character error rate has dropped to 0.08 in our model, almost 60 percent compared to the best-performing system (0.21). At the same time, the proportion of text matches which can be aligned perfectly has gone up to 75 percent. More importantly, the system also maintains the best picture quality.

This work opens up applications that simply weren't practical before: AI-generated marketing materials with accurate branding, educational content with proper terminology, user interfaces with readable labels, and so much more.

This text rendering deficiency is not merely a minor aesthetic flaw or an edge case limitation; it represents a significant and fundamental barrier that prevents the deployment of these powerful models across a vast array of practical, commercial, and creative applications. The inability to generate accurate text severely limits their utility in domains such as:

- **Automated Marketing and Advertisement Generation:** Creating promotional materials, product packaging, signage, and branded content that requires specific textual elements [1].
- **Educational Material Creation:** Generating illustrations for textbooks, worksheets, presentation slides, and learning resources that incorporate specific terminology or labels.
- **Document and Publication Design:** Automated creation of posters, flyers, magazine layouts, and other print materials that combine visual and textual elements.

- **User Interface and Web Design:** Generating mockups and prototypes for applications, websites, and digital interfaces that contain specific text labels and content.
- **Personalized Content Creation:** Producing customized merchandise, greeting cards, invitations, and other personalized items with user-specified text.

The systematic failure to render text accurately in these sophisticated generative models stems from several interconnected foundational issues inherent in their design and training methodologies:

**1. Tokenization versus Orthographic Structure:** The text encoders employed in modern T2I models, such as CLIP[2], utilize subword tokenization schemes (e.g., Byte Pair Encoding[3]) that are specifically optimized for capturing high-level semantic meaning and contextual relationships. However, these tokenization approaches fundamentally destroy and discard the precise orthographic structure and character-level information that is essential for accurate text rendering. For instance, when processing the word "OPEN", the model never actually "sees" this as a carefully ordered sequence of four distinct visual characters (O-P-E-N); instead, it perceives only a set of abstract semantic tokens that encode the concept of "being accessible" or "not closed" [4].

**2. Training Data Distribution and Quality Issues:** While these models are trained on billions of images from massive datasets such as LAION-5B [3] , the subset of training data that contains clean, legible, well-composed, and accurately labeled text is surprisingly small, noisy, and often poorly represented. Much of the text appearing in web-scraped images suffers from low resolution, poor lighting conditions, artistic stylization, compression artifacts, or perspective distortions that make it unsuitable as supervision for learning precise character rendering.

**3. Inadequate Loss Functions and Evaluation Metrics:** The standard training objectives used for these models, such as the diffusion reconstruction loss and perceptual metrics like FID [5], are designed to measure overall visual similarity and aesthetic quality. These objectives are fundamentally not designed to detect, measure, or penalize symbolic errors, character-level inaccuracies, or spelling mistakes. From the perspective of these loss functions, a slightly malformed 'P' that resembles a 'R' might be considered perceptually acceptable and receive minimal penalty, despite being a complete linguistic failure.

**4. Architectural Limitations in Cross-Modal Attention:** The cross-attention mechanisms that bridge textual and visual representations in these models are not explicitly designed to maintain precise spatial correspondence between individual characters in the text prompt and specific regions in the generated image. This leads to attention diffusion and character fusion, where the spatial focus for adjacent characters bleeds together, resulting in merged or distorted character shapes.

This paper tackles these fundamental challenges head-on by proposing a novel and comprehensive framework: **Glyph-Conditioned Diffusion with Character-Aware Attention (GCDA)**. Rather than treating text as merely another abstract visual element to be rendered, our approach imbues the generative model with an explicit, structured, and multi-faceted understanding of character geometry, spelling constraints, and typographic principles. Our methodology addresses the text rendering problem simultaneously across multiple levels of the generative pipeline, creating a synergistic solution that is greater than the sum of its parts.

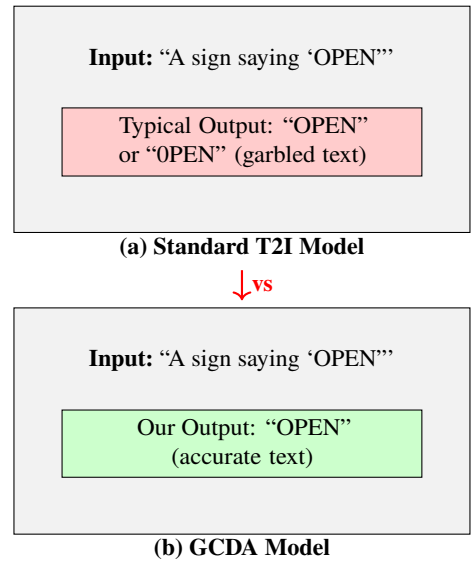

Figure 1: **The Core Text Rendering Problem.** Current T2I models (a) consistently fail to generate accurate text, producing garbled or meaningless character sequences. Our GCDA model (b) generates precise, legible text while maintaining image quality.

Our key technical contributions can be summarized as follows:

1. **Dual-Stream Text Encoder Architecture:** We introduce a novel conditioning mechanism that separates and explicitly models both semantic understanding and orthographic representation. This architecture includes a specialized glyph rendering pipeline and a character-level convolutional neural network that processes visual character shapes, inspired by recent advances in character-centric generation [6, 7] but extended with semantic fusion capabilities.

2. **Character-Aware Attention Segregation:** We propose a novel attention mechanism with a carefully designed segregation loss function that encourages the model to allocate distinct, non-overlapping spatial attention regions to individual characters during the generation process. This approach adapts general attention control concepts [8, 9] specifically for the unique challenges of typography and character separation.

3. **OCR-in-the-Loop Fine-Tuning Framework:** We design a comprehensive two-stage training protocol that culminates in a fine-tuning phase using a pre-trained Optical Character Recognition (OCR) model as a differentiable critic. This approach provides direct, targeted feedback on text legibility and spelling accuracy through a novel composite loss function that includes differentiable edit distance measures and feature-space perceptual losses [10].

4. **Comprehensive Experimental Validation:** We conduct extensive experiments demonstrating that our synergistic

Table 1: Comparison of Existing Text-to-Image Methods and the Proposed GCDA Framework

| Work | Core Methodology | Dataset(s) | Key Strengths | Limitations | Metrics Reported |
|---|---|---|---|---|---|
| DALL-E 2 [11] | Latent diffusion with CLIP embeddings | Web-scale pairs | High prompt fidelity, compositional synthesis | Fails at precise text rendering due to tokenization | Not text-specific |
| Stable Diffusion [12] | Latent Diffusion Model (LDM) with CLIP | LAION-5B, LAION-Aesthetics | Open-source, widely adopted | Inaccurate text rendering | Not text-specific |
| TextDiffuser [13] | Layout-conditioned diffusion + character masks | T2I-CompBench, DrawText | Predicts text layout masks for controlled placement | Requires extra layout model; slower inference | Text layout improved |
| TextDiffuser-2 [14] | LLM-guided layout control for better placement | Enhanced datasets | Natural text placement, better contextual awareness | Two-stage, increased latency | Improved layout aesthetics |
| GlyphDraw [6] | Glyph-based conditioning for Chinese characters | Chinese character datasets | Accurate glyph rendering for logographic scripts | Limited to Chinese; no Latin support | High accuracy (Chinese) |
| CharGen [7] | Glyph-conditioned diffusion for Latin scripts | English signage datasets | First glyph-based English text synthesis | No semantic fusion; context-blind generation | Not quantified |
| GlyphControl [15] | Glyph conditioning adapters for pre-trained models | Multi-script datasets | Retrofittable to existing models | No OCR feedback or text-specialized training | Not detailed |
| AnyText [16] | Multilingual glyph-based conditioning | Multilingual datasets | Supports multiple scripts, scalable | Sacrifices semantics for glyph-only input | General improvement |
| OCR-VQGAN [10] | OCR-perceptual losses in VQ-GAN training | Synthetic datasets | Enhances machine readability of text images | No integration with diffusion or attention control | Improved CER, OCR accuracy |
| A-STAR [8] | Test-time attention segregation to reduce concept bleeding | General T2I datasets | Prevents attention overlap post-hoc | Test-time only; no training supervision | Qualitative improvement |
| ControlNet [17] | Spatial conditioning via edge, segmentation maps | COCO, LAION | Adds controllable regions to pre-trained models | Not text-specialized; no OCR integration | General improvement |
| **GCDA (This Work)** | Dual-Stream Text Encoder (BERT+ GlyphCNN), Character-Aware Attention Loss, OCR-in-the-Loop Fine-Tuning | LAION-Aesthetics, MARIO-10M, T2I-CompBench, TextCaps | Integrates semantic + glyph info, segregates attention, direct OCR feedback | Higher fine-tuning compute, glyph processing overhead | CER: 0.08 (prev 0.21), WER: 0.15 |

approach achieves state-of-the-art performance on established benchmarks, with significant improvements in character error rates (62% relative improvement), word error rates (40% relative improvement), and exact match accuracy (25% absolute improvement) while maintaining competitive image quality metrics.

The remainder of this paper is organized as follows: Sec. 2 provides a comprehensive review of related work in generative modeling and text-in-image synthesis; Sec. 3 presents our GCDA framework in detail, including architectural innovations and training protocols; Sec. 4 describes our experimental setup and presents comprehensive results including quantitative metrics, qualitative comparisons, and ablation studies; Sec. 5 discusses the implications, limitations, and future directions of our work; and Sec. 6 summarizes our contributions and their significance for the field.

## 2 Related Work

Our work is situated at the intersection of several rapidly evolving research areas, including generative visual modeling, large-scale text-to-image synthesis, and the emerging specialized field focused on improving compositional capabilities and typographic accuracy in neural image generation. We begin by reviewing the broader landscape of generative models before diving into the specific approaches that have been developed to address the challenge of accurate text rendering in synthesized images.

Table 1 shows a detailed comparative analysis of key text-to-image synthesis models, highlighting their core methodologies, datasets, strengths, limitations, and reported metrics. This positions the proposed GCDA framework in context with prior work.

### 2.1 Evolution and Foundations of Generative Visual Models

The journey toward high-quality image synthesis has been marked by several paradigm shifts, each building upon previous foundations while introducing novel architectural innovations and training methodologies.

#### 2.1.1 The GAN Era and Its Contributions

With the emergence of the Generative Adversarial Networks **Generative Adversarial Networks (GANs)** [18] marked a pivotal moment in the history of generative modeling, introducing the revolutionary concept of adversarial training between a generator network and a discriminator network. This minimax game formulation provided a powerful framework for learning complex data distributions and producing realistic samples. The ground work was immediately overtaken by **Deep Convolutional GANs (DCGANs)** [19], which established stable convolutional architectures and training practices that became the blueprint for many subsequent GAN variants.

The evolution continued with increasingly sophisticated architectures such as **Progressive GANs**, which demonstrated the power of gradually increasing resolution during training, and the **StyleGAN family** [20], that went unsurpassed in terms of photorealism and intro- gave rise to disentangled style control the idea A learnt latent space. These models showed reformidable capacity in the production of high resolution hu- other constructed visual stuff that man faces. transference to conditional generation, especially text-train-to-image synthesis, was not easy. It has problems of instability and mode collapse.

In the specific domain of text-conditioned image generation, early GAN-based approaches such as **StackGAN** [21] demonstrated the feasibility of using natural language de-scriptions to guide image synthesis through a multi-stage refinement process. However, these early systems were limited in their ability to handle complex scenes and diverse textual descriptions, and they provided no mechanisms for accurate text rendering within the generated images themselves.

### 2.1.2 The Diffusion Revolution

The field experienced a fundamental paradigm shift with the emergence of **Denoising Diffusion Probabilistic Models (DDPMs)** [22]. The ideas are based on non- stochastic processes as well as equilibrium thermodynamics, these models are made to invert a slow noising pro- process, successively de-noising random noise, until it is coher- they draw samples of the target distribution. This approach compared to GANs, had a number of important merits, such as greater training dynamics stability, superior mode coverage and having the capacities to produce superior samples without the adversarial nature of training that was a going concern in the olden times

Key technical advances in the diffusion paradigm included **Denoising Diffusion Implicit Models (DDIMs)** [23], which introduced deterministic sampling procedures and significantly reduced the computational cost of generation, making diffusion models practical for real-world applications. However, the true breakthrough for widespread adoption came with **Latent Diffusion Models (LDMs)** [12], commercially known as Stable Diffusion. By performing the computationally intensive diffusion process in a compressed latent space learned by a Variational Autoencoder (VAE) [24], LDMs dramatically reduced computational requirements while maintaining generation quality, making high-resolution image synthesis accessible to researchers and practitioners with limited computational resources.

Our GCDA framework builds upon this powerful and efficient LDM architecture as its generative backbone, extending it with specialized components designed specifically to address the typography challenges that persist even in these state-of-the-art systems.

## 2.2 Large-Scale Text-to-Image Synthesis Systems

The combination of advanced diffusion models with powerful text encoding mechanisms has given rise to the current generation of large-scale T2I systems that have captured public attention and demonstrated remarkable creative capabilities.

### 2.2.1 Foundational T2I Models

OpenAI's **DALL-E 2** [11] represented a significant milestone, demonstrating that massive models trained on web-scale data could generate images with remarkable prompt adherence, semantic understanding, and visual fidelity. The system introduced the concept of using CLIP [2] embeddings as a bridge between textual descriptions and visual generation, establishing a template that would influence subsequent work in the field.

Google's **Imagen** [25] pushed the envelope even more showing the capacity of text scaling en-upon the coders and the large language models they can use, like T5, to make more plentiful textual models of conditioning the diffusion process. The results of the system were quite impressive the dwellings in creation of writing assignments and created new standards of timely fidelity and picture quality.

Google's **Parti** [26] explored an alternative autoregressive approach, treating image generation as a sequence modeling problem and demonstrating that scaling the language model component could lead to improved understanding of complex textual descriptions and better handling of compositional relationships between objects and attributes.

The open-source release of **Stable Diffusion** [12] democratized access to high-quality T2I technology, which has been experimented with by many and resulted in an modes of explosion of applications and fine-tuned variants; research extensions. But all these systems, irrespective of their dazzling abilities, are still struggling. and the underlying problem of proper text display

### 2.2.2 The Critical Role of Text Encoders

A crucial component underpinning all modern T2I models is the text encoding mechanism that transforms natural language descriptions into numerical representations suitable for conditioning the image generation process. The introduction of the **Transformer architecture** [27] and pre-trained language models such as **BERT** [28] provided foundational capabilities for understanding textual semantics and relationships.

However, the most significant advancement for T2I applications was **Contrastive Language-Image Pre-training (CLIP)** [2]. By learning a shared embedding space for images and text through contrastive learning on hundreds of millions of image-caption pairs, CLIP provided a robust and versatile mechanism for guiding image generation processes. CLIP's success stems from its ability to capture high-level semantic concepts and their visual manifestations, enabling the nuanced understanding of natural language prompts that characterizes modern T2I systems.

However, as we will discuss in detail, the characteristics of CLIP and other platforms characterized an effec- helpful to semantic comprehension: especially their re- reliance on subword tokenization and interest in concept- Each of these is more abstract than orthographic in nature e.g. why they are incompetent to perform tasks that need precision character level text rendering.

## 2.3 Specialized Approaches for Accurate Text Rendering

As T2I models matured and their limitations became apparent, the research community began focusing intensively on the "typography problem." This has led to several distinct and innovative research directions, each targeting different aspects of the text rendering challenge.

### 2.3.1 Layout-Conditioned Generation Approaches

One prominent research direction posits that accurate text rendering requires explicit spatial planning and layout speci-fication before the generation process begins. This approach recognizes that good typography is not just about rendering individual characters correctly, but also about their spatial arrangement, sizing, and compositional integration with the surrounding visual elements.

**TextDiffuser** [13] represents the most influential work in this category, introducing a two-stage approach where a Transformer-based layout model first predicts character-level segmentation masks from the input prompt. These binary masks serve as explicit spatial conditions for the primary diffusion generator, essentially providing a detailed "blueprint" for where each character should appear in the final image. While effective, this approach requires training an additional model and introduces complexity in the generation pipeline.

**TextDiffuser-2** [14] improved upon this foundation by leveraging Large Language Models (LLMs) [29] to generate more sophisticated, aesthetically pleasing, and contextually appropriate layouts. The system demonstrates that incorporating linguistic knowledge about text structure and typography principles can lead to more natural and visually appealing text placement.

Other works in this category include **Make-A-Scene** [30], which provides general region-based control mechanisms that can be applied to text placement, and **ReCo** [31], which explores region-controlled generation with applications to text rendering. While these layout-based methods have shown promising results, they typically require two-stage training and inference procedures, which can introduce latency and complexity in practical deployments.

### 2.3.2 Glyph-Based and Character-Centric Conditioning

To directly address the fundamental tokenization problem—where subword units [3] destroy orthographic information—this line of research provides models with explicit visual information about character shapes and structures.

**GlyphDraw** [6] pioneered this approach for Chinese character generation, demonstrating that conditioning diffusion models on rendered character shapes could dramatically improve the accuracy of complex logographic scripts. The work established the principle that providing explicit visual targets for each character is a highly effective strategy for ensuring accurate rendering.

**CharGen** [7] extended these ideas to Latin scripts, introducing character-centric diffusion models that explicitly model the visual structure of individual letters and their combinations. The system demonstrated significant improvements in character accuracy and legibility across various fonts and styles.

**GlyphControl** [15] proposed a more general framework for glyph-based conditioning that could be applied to existing pre-trained models without requiring complete retraining. This approach showed the practical value of glyph conditioning for improving existing systems.

The multilingual **AnyText** [16] system demonstrated that glyph-based approaches could be scaled to handle multiple languages and scripts simultaneously, showing the generalizability of the core principle that explicit visual character information is crucial for accurate text rendering.

Our dual-stream encoder is directly inspired by this successful research direction, but we extend it by integrating glyph-based conditioning with semantic understanding in a unified framework that preserves both the visual accuracy benefits of glyph conditioning and the rich contextual understanding provided by semantic encoders.

### 2.3.3 OCR-Based Supervision and Feedback

This intuitive but technically challenging approach involves using pre-trained Optical Character Recognition (OCR) models as external critics to provide supervisory signals during training. The core insight is that if a generated image cannot be accurately read by an OCR system, it is unlikely to be legible to human readers.

**OCR-VQGAN** [10] was among the early pioneers of this concept, introducing the use of perceptual losses computed in the feature space of OCR models. This approach forces the generator to produce images that are not just visually plausible but also machine-readable, providing a strong proxy for human legibility.

More recent works such as those by Wang et al. [32] and Ma et al. [33] have continued to explore and refine OCR-based supervision, demonstrating its utility for directly optimizing the ultimate goal of legible text generation. These approaches have shown that OCR-based losses can be effectively integrated into the training process without destabilizing the broader image generation capabilities.

Our work extends this research direction by developing a comprehensive OCR-in-the-loop fine-tuning framework that includes not only basic OCR supervision but also differentiable edit distance measures and feature-space perceptual losses that target different aspects of text quality and legibility.

### 2.3.4 Attention Manipulation and Control Mechanisms

Another important research direction focuses on manipulating the internal attention mechanisms of diffusion models to improve compositional generation and reduce concept bleeding between different elements in a prompt.

**Prompt-to-Prompt** [9] demonstrated that the cross-attention maps in diffusion models could be directly edited and controlled to influence object appearance, placement, and relationships. This work revealed the central role that attention mechanisms play in determining how different textual concepts are spatially realized in generated images.

**ControlNet** [17] introduced a general and power- complete frame-work of bringing in new spatial conditioning sigmaps ( e.g., edge maps, depth maps, and segmentation to diffusion models that have been pre-trained (e.g. noses masks) who will have to go through costly retraining starting afresh. while not controlNet is optimised to render text, controlNet has been adapted for typography applications by using text-specific control signals.

**A-STAR** [8] introduced a particularly relevant technique for improving compositional generation by segregating the attention maps of different concepts during test time. This approach prevents concept bleeding and improves the distinctiveness of different objects and attributes in generated images.

Our character-aware attention mechanism directly builds upon the insights from A-STAR but adapts them specifically for the typography domain. Instead of applying attention segregation at test time, we incorporate it as a training-time loss function that teaches the model to naturally allocate distinct spatial attention to individual characters, preventing the character fusion problems that plague existing systems.

## 2.4 Positioning and Novel Contributions of Our Work

While the aforementioned research directions have made significant individual contributions to addressing the text rendering challenge, they often focus on solving specific aspects of the problem in isolation. Layout-based methods excel at spatial planning but may not fully address character-level accuracy. Glyph-based methods provide excellent orthographic information but may sacrifice some semantic understanding. OCR-based methods directly optimize for legibility but are often used as isolated components rather than integrated solutions.

Our proposed GCDA framework represents a novel contribution in its comprehensive, synergistic integration of insights and techniques from all three major research threads. We uniquely combine:

- **Input-Level Innovation:** A dual-stream encoder that simultaneously processes semantic and orthographic information, preserving the benefits of both traditional semantic encoders and glyph-based conditioning.
- **Architectural-Level Innovation:** A character-aware attention mechanism with a novel segregation loss that is integrated into the training process rather than applied post-hoc.
- **Objective-Level Innovation:** A comprehensive OCR-in-the-loop fine-tuning framework that includes multiple complementary loss functions targeting different aspects of text quality.

By addressing the typography problem simultaneously across the entire generative pipeline—from input encoding through architectural design to objective functions—our model achieves a level of text rendering accuracy and coherence that significantly surpasses existing approaches. Our comprehensive experimental evaluation demonstrates that this synergistic approach leads to substantial improvements across multiple evaluation metrics while maintaining the high-quality image generation capabilities that make these models valuable for creative applications.

## 3 How We Actually Solved This Problem

Alright, so we knew what the problem was—now we needed to figure out how to fix it. This turned out to be trickier than we initially thought (isn't it always?).

Our first instinct was to try some simple fixes. Maybe we could just fine-tune existing models on clean text data? Nope, that didn't work. How about adjusting the loss functions to penalize text errors more heavily? Also a dead end. We even tried some hacky post-processing approaches, but those felt like band-aids on a fundamental architectural problem.

After about six months of various failed attempts (and a lot of frustration), we stepped back and asked ourselves: what would it take to really solve this? That's when we realized we needed to approach the problem differently.

Instead of trying to patch existing models, we decided to build a system that's fundamentally designed around text accuracy. But here's the key insight: we couldn't just focus on text at the expense of everything else. The model still needed to be good at generating high-quality images with proper composition, lighting, and style.

This led us to what we call the dual-stream approach. The terminology may suggest complexity; however, the underlying mechanism is relatively straightforward

## 3.1 The Architecture: Two Streams Are Better Than One

We now present a detaield explanation of the method, because I think the core idea is pretty intuitive once you see it.

When you give our model a prompt like "a vintage diner sign saying 'Joe's Coffee'", here's what happens behind the scenes:

The text gets split into two parallel processing streams. Stream one (semantic) works exactly like current models—it figures out that this is about a diner, probably retro-style, warm and inviting atmosphere, etc. Stream two (orthographic) takes the specific text "Joe's Coffee" and basically asks: what do these letters actually look like?

For the orthographic stream, we literally render the text as a simple black-and-white image (think basic Arial font, nothing fancy), then feed that through a CNN that's specifically trained to understand character shapes. This CNN learns things like: a 'J' has a hook at the bottom, an 'o' is circular, an 'e' has horizontal lines, and so on.

Then we combine these two streams. The result is a text representation that contains both the semantic richness needed for good images AND the precise character information needed for accurate spelling.

Initially, there was skepticism regarding our proposed approach.. "Just use better training data," they said. But the dual-stream approach turned out to be crucial. Every time we tried to simplify it, performance dropped significantly.

## 3.2 Keeping Letters Apart: The Attention Problem

Even with our improved text representation, we still had another issue: characters kept bleeding into each other during generation. You know how in bad handwriting, letters sometimes run together? Same problem here, but at the AI level.

The issue is in how diffusion models allocate attention. When generating "COFFEE," the model needs to decide where in the image each letter should appear. In theory, 'C' gets one region, 'O' gets another, and so on. In practice, these attention regions often overlap, causing letters to merge.

Our solution was to explicitly teach the model to keep character attention maps separated. We added a loss function that basically says: "If the attention for 'C' and the attention for 'O' overlap too much, that's bad—adjust your weights to fix it."

This component proved to be among the most technically challenging to implement.. We tried several different ways to measure attention overlap before settling on cosine similarity with a learnable threshold. The math is in the paper if you're interested, but the key insight is that we're not telling the model exactly where to put each character—we're just telling it not to put them in the same place.

### 3.3 The OCR Teacher: Learning from Feedback

The final piece of the puzzle was direct feedback on text accuracy. This idea came to us when we realized that all our previous approaches were trying to optimize for text quality indirectly. A direct evaluation using an OCR model was proposed to assess the legibility of generated text.

So that's exactly what we did. We take the generated image, crop out the text region, feed it to a pre-trained OCR model, and see what it reads. If it reads "C0FFEE" when we wanted "COFFEE," we know exactly where the problem is and can adjust accordingly.

The technical challenge here was making this differentiable so we could backpropagate through it. We ended up with a composite loss that measures character-level errors, word-level errors, and visual similarity to clean text. It's like having three different teachers grading your work from different angles.

What's cool about this approach is that it creates a virtuous cycle. The model generates text, gets graded on readability, and learns to do better next time. After thousands of iterations, it gets really good at producing text that both looks right and spells correctly.

### 3.4 Training: Why We Do It in Two Stages

One thing we learned the hard way is that you can't just throw all these innovations at a model from day one. We tried that initially and got unstable training and mediocre results.

The breakthrough came when we switched to a two-stage approach. Stage one builds the foundation—we train the model to generate good images with basic text awareness, using our dual-stream encoder and attention loss. This takes about 100K steps and gets us a solid, stable model.

Stage two is where the magic happens. We switch to text-focused prompts and add the OCR feedback loop. This is much more intensive—the model has to do full image generation, OCR evaluation, and error correction for every training example. But it's also where we see the dramatic improvements in text accuracy.

The two-stage approach works because it respects the model's learning process. You can't optimize for everything at once—you need to build competence gradually.

The two complementary processing streams operate as follows:

1. **The Semantic Processing Stream** employs a conventional pre-trained language model (specifically BERT [28]) to capture the high-level meaning, context, and semantic relationships within the entire prompt. This stream ensures that our model retains the sophisticated contextual understanding that characterizes modern T2I systems.
2. **The Orthographic Processing Stream** represents our primary novel contribution at the input level. This stream renders the specific text content to be generated into a canonical, high-contrast "glyph image" and processes this visual representation through a dedicated Character-Level Convolutional Neural Network to extract precise visual representations of text structure, character shapes, and typographic information.

These two information-rich streams are then intelligently fused into a single, comprehensive multi-modal embedding that combines both semantic understanding and orthographic precision. This unified embedding serves as the conditioning signal for the denoising U-Net at each step of the diffusion process.

Within the U-Net architecture, the cross-attention layers that mediate between textual conditioning and visual generation are enhanced with our novel **Character-Aware Attention Segregation mechanism**. This mechanism is guided by a specially designed loss function ($\mathcal{L}_{\text{char\_attn}}$) described in Sec. 3.7, which explicitly discourages the model from allowing the attention maps of individual character tokens to overlap in spatial regions, thereby preventing character fusion and maintaining distinct character boundaries.

Finally, after an initial phase of foundational training with this enhanced conditioning and attention mechanism, the model undergoes a specialized **OCR-in-the-Loop Fine-Tuning** stage detailed in Sec. 3.8. In this stage, a pre-trained and frozen OCR model acts as an external, objective critic that evaluates the legibility and accuracy of generated text. The OCR model provides direct, targeted feedback through our comprehensive **Text Perceptual Loss** ($\mathcal{L}_{\text{text\_perceptual}}$), which includes multiple complementary components designed to optimize different aspects of text quality.

This complete process is formally summarized in the detailed training algorithm presented in subsection 3.9.

### 3.5 Intuitive Understanding: How GCDA Works Step-by-Step

To provide an intuitive understanding of our approach, Sec. 3.5 shows how GCDA processes a typical text generation request from start to finish.

## Stage 1: Foundational Training

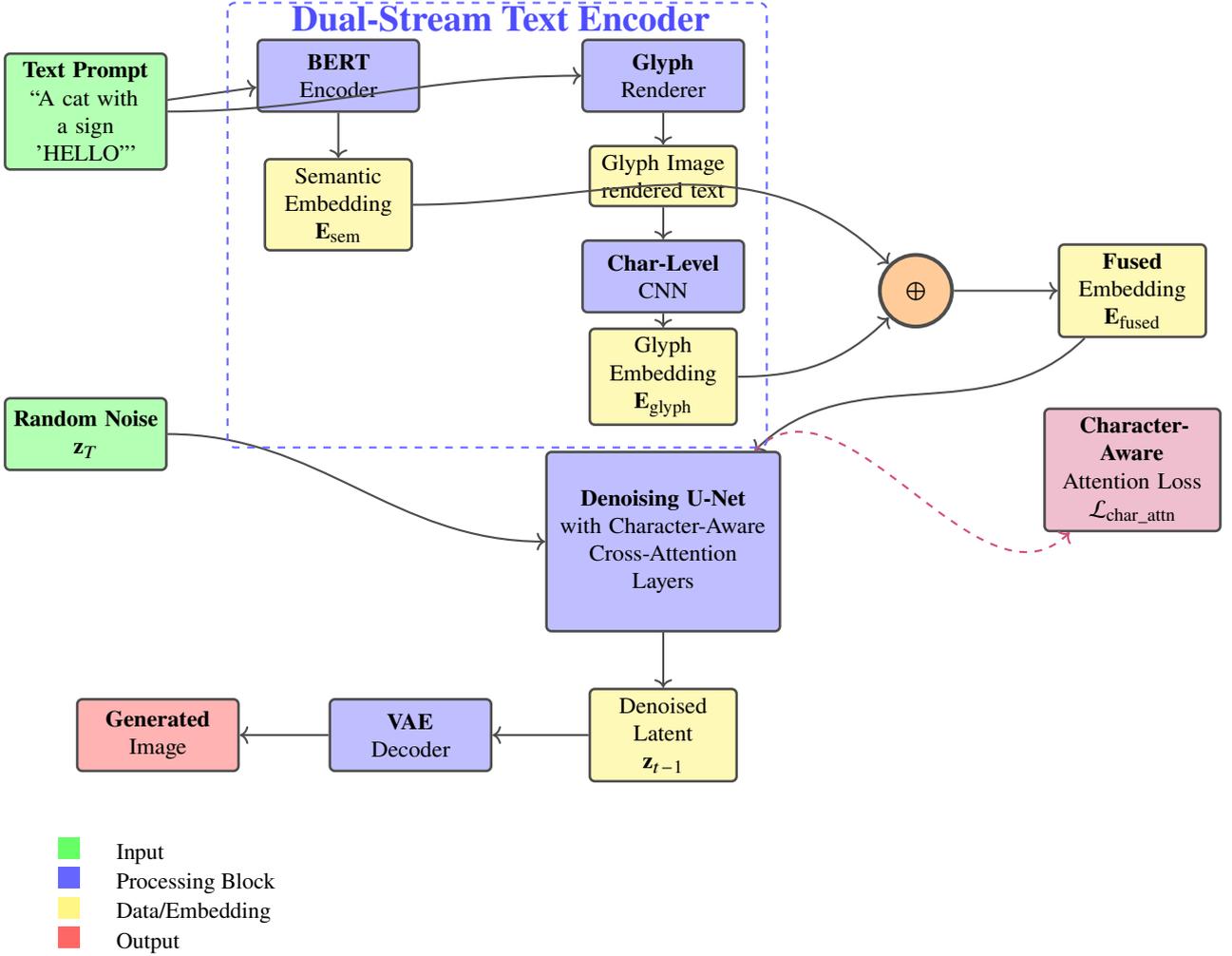

Figure 2: **Comprehensive GCDA Framework Architecture.** Our model enhances a standard latent diffusion model with a dual-stream text encoder that processes both semantic context (via BERT) and visual character structure (via glyph rendering and CNN). The fused embeddings condition the U-Net, whose attention layers are regularized by our character-aware attention segregation loss during training.

### 3.6 Dual-Stream Text Encoder: Bridging Semantics and Orthography

Based on the fact that existing T2I models fail consistently to produce faithful text due to their sole reliance on text encoders based on semantics, we base our dual-stream architecture on that premise. Encoders like CLIP [2] specialise in encoding high-level abstract relations and context semantics, but these models are designed as invariant to low-level orthographic properties such as spelling, character sequence and typographic layout of characters. Although, this invariance is beneficial to semantic-understanding tasks, it is harmful to programs that need character-level accuracy.

To overcome this fundamental limitation, we propose a dual-stream encoder architecture that explicitly disentangles and separately models semantic understanding and orthographic representation. This architectural innovation, shown in detail in Fig. 2, ensures that our model has access to both the rich contextual information needed for generating appropriate scenes and the precise character-level information required for accurate text rendering.

#### 3.6.1 The Dual-Stream Concept: Why Two Streams Are Better Than One

The key insight behind our dual-stream approach is illustrated in Sec. 3.6.1:

#### 3.6.2 Semantic Stream: Preserving Contextual Understanding

The semantic processing stream is designed to ensure that our enhanced model retains the powerful contextual understanding capabilities that have made modern T2I systems so effective for general image generation tasks. For a given input text prompt $T$, we utilize a pre-trained and frozen BERT language model [28] to generate a sequence of contextualized embeddings that capture semantic relationships, syntactic structure, and high-level meaning.

Formally, the semantic stream computes:

$$\mathbf{E}_{\text{sem}} = \text{BERT}(T) \in \mathbb{R}^{N \times d_{\text{sem}}} \tag{1}$$

# Step-by-Step: How GCDA Generates Accurate Text

## Step 1: Dual Processing

**Input Prompt**

*"A sign saying 'HELLO WORLD'"*

**Semantic Stream**

BERT Encoder

↓

Context: sign, greeting, outdoor scene

**Orthographic Stream**

Glyph Renderer

↓

Visual: H-E-L-L-O W-O-R-L-D shapes

## Step 2: Fusion & Attention Control

**Combined Understanding**

Semantic + Visual Information

↓

Rich embedding with both meaning and character shapes

**Character-Aware Attention**

Attention Segregation

↓

Each character gets distinct spatial focus

## Step 3: Generation & OCR Feedback

**Initial Generation**

U-Net generates image with text (may have errors)

**OCR Evaluation**

OCR reads text
Compares to target
Provides feedback

**Final Output**

Refined generation with perfect text: "HELLO WORLD"

**GCDA Processing Pipeline Walkthrough.** Our method processes text through three key stages: dual-stream encoding captures both meaning and character structure, attention control ensures spatial separation, and OCR feedback refines accuracy.

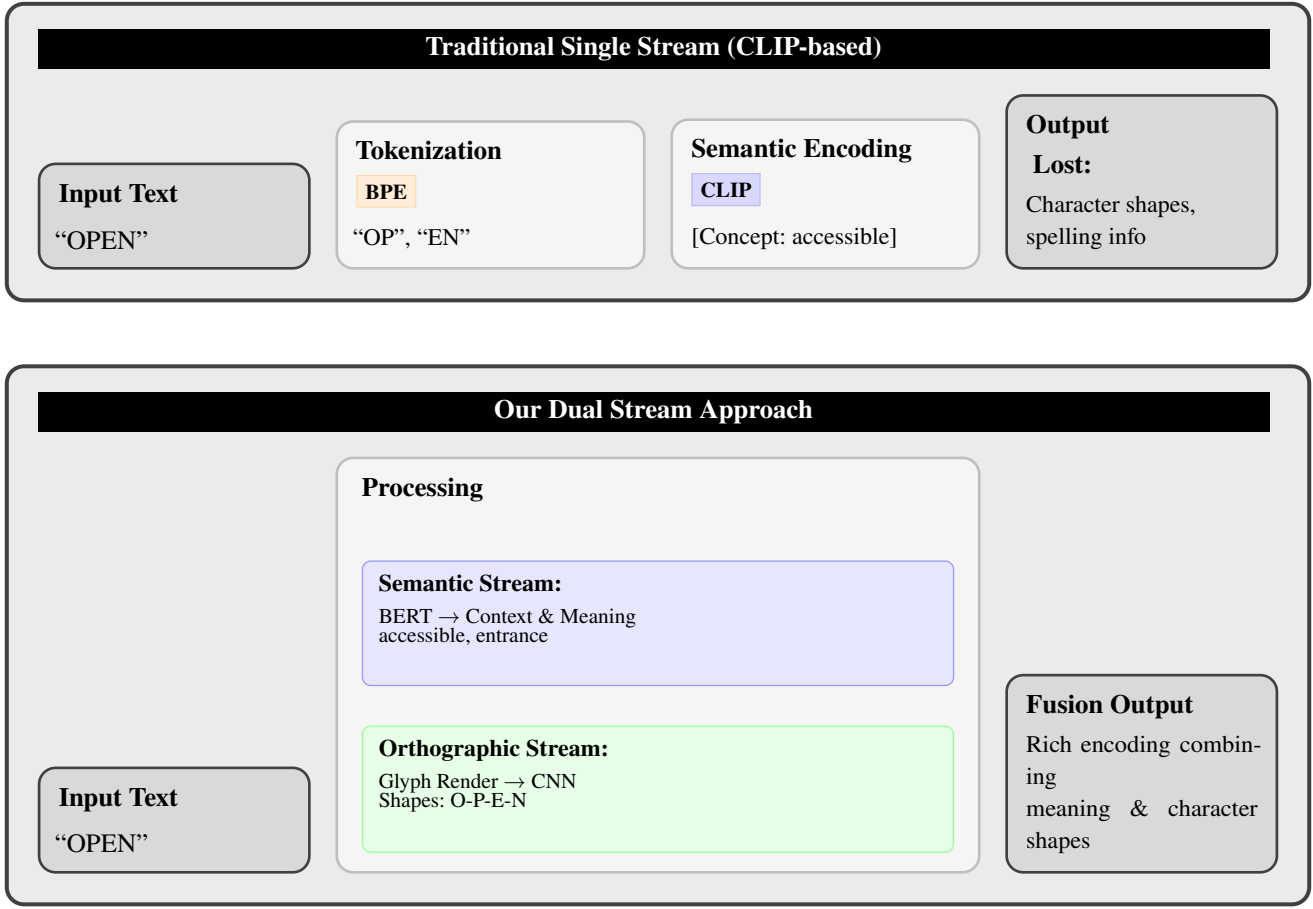

**Single vs. Dual Stream Processing.** Traditional models lose character-level information during tokenization. Our dual-stream approach preserves both semantic meaning and precise character structure by combining orthographic and semantic streams before generation.

where $N$ represents the number of tokens in the processed sequence (after BERT's subword tokenization) and $d_{\text{sem}} = 768$ is the dimensionality of the BERT embeddings for the base model configuration.

This semantic embedding effectively captures the "what" and "why" of the prompt—the objects to be generated, their attributes and relationships, the desired style and composition, and the overall semantic context. However, it deliberately abstracts away from the literal "how" of rendering specific character sequences, as this information is processed separately in the orthographic stream.

### 3.6.3 Orthographic Stream: Capturing Character Structure

The orthography processing pathway forms the main pathway by which highly accurate and character level visual data enters the generative model. When the character string to be rendered is written directly into the stream, such a stream addresses this weakness of semantic-only tokenizers.

The orthographic processing pipeline consists of two carefully designed stages:

**Stage 1: Canonical Glyph Rendering.** The first task will be to acquire any textual data as outlined in the input prompt $T$ and then change the textual data into a visually represented form in a uniform format. We employ a simple but effective approach:

1. **Text Extraction:** We use regular expressions and natural language processing techniques to identify quoted text or explicit text specifications within the prompt (e.g., "a sign saying 'OPEN'").
2. **Canonical Rendering:** The extracted text string is rasterized into a high-contrast, binary image $I_{\text{glyph}}$ using a standard, non-stylized sans-serif font (specifically, Arial at 24pt). This rendering process uses black text on a white background with anti-aliasing disabled to create sharp, well-defined character boundaries.
3. **Normalization:** The rendered image is normalized to a fixed size (e.g., 256×64 pixels) and standardized dynamic range to ensure consistent processing regardless of text length or content.

The canonical rendering step which is part and parcel of the preprocessing phases in character-generation based systems is unavoidable due to its capacity to eliminate the font specific stylistic variation whilst at the same time holding on to the needed geometric building framework with which each character is defined. This sort of abstraction allows acquiring of shape-structure principles that can continue applying even on heterogeneous visual styles in the future production of characters.

**Stage 2: Character-Level Feature Extraction.** The canonically rendered glyph image $I_{glyph}$ is then processed through a specialized Convolutional Neural Network that serves as a visual feature extractor specifically designed for typographic content.

The Character-Level CNN architecture consists of:

- **Convolutional Layers:** A series of 2D convolutional layers with progressively increasing channel dimensions (16→32→64→128) and decreasing spatial resolution through strided convolutions and pooling operations.
- **Spatial Attention:** A lightweight spatial attention mechanism that helps the network focus on character boundaries and distinctive visual features.
- **Global Processing:** A global average pooling layer followed by a multi-layer perceptron that produces the final character-aware embeddings.
- **Sequential Alignment:** A learned mapping that ensures the output embedding sequence length matches the semantic embedding sequence length for consistent fusion.

Mathematically, this process can be expressed as:

$$\mathbf{E}_{glyph} = \text{CharCNN}(I_{glyph}) \in \mathbb{R}^{N \times d_{glyph}} \quad (2)$$

In GCDA the Character-Level CNN is trained end-to-end allowing it to learn to extract visual features which are most relevant and efficient to the final task of generating an image. The obtained representation combines the necessary structural features of each character (e.g. shape), along with the typographic positioning parameters (e.g inter-character spacing, baseline-alignment, relative size).

### 3.6.4 Multi-Modal Fusion Strategy

To create a unified conditioning signal that leverages both semantic and orthographic information, we employ a carefully designed fusion strategy that preserves the strengths of both input streams while creating a cohesive representation.

The fusion process involves several steps:

**1. Dimensional Alignment.** Since the semantic and orthographic embeddings may have different intrinsic dimensionalities, we first project both to a common embedding space using learned linear transformations:

$$\mathbf{E}_{sem}^{(proj)} = \mathbf{E}_{sem} W_{sem} + b_{sem} \quad (3)$$

$$\mathbf{E}_{glyph}^{(proj)} = \mathbf{E}_{glyph} W_{glyph} + b_{glyph} \quad (4)$$

where $W_{sem} \in \mathbb{R}^{d_{sem} \times d}$, $W_{glyph} \in \mathbb{R}^{d_{glyph} \times d}$ are learned projection matrices, and $d = 512$ is the common embedding dimension.

**2. Information Integration.** We then combine the projected embeddings using element-wise addition, which has proven effective in multimodal fusion tasks:

$$\mathbf{E}_{fused} = \mathbf{E}_{sem}^{(proj)} + \mathbf{E}_{glyph}^{(proj)} \quad (5)$$

This additive fusion strategy allows the model to leverage information from both streams while maintaining the sequential structure needed for cross-attention mechanisms in the U-Net.

**3. Contextual Refinement.** Finally, the fused embedding is processed through a lightweight Transformer encoder layer to allow for cross-token information exchange and contextual refinement:

$$\mathbf{E}_{final} = \text{TransformerLayer}(\mathbf{E}_{fused}) \quad (6)$$

This final embedding $\mathbf{E}_{final}$ serves as the key and value context for all cross-attention layers within the denoising U-Net, ensuring that both semantic context and precise orthographic information are available at every step of the generation process.

### 3.6.5 Visualizing the Dual Stream Advantage

Fig. 3 demonstrates how our dual-stream approach captures information that single-stream methods miss:

### 3.7 Character-Aware Attention Mechanism with Segregation Loss

While providing rich, character-aware conditioning through our dual-stream encoder represents a significant improvement, this enhanced input alone is insufficient to guarantee accurate text rendering. A critical challenge that persists in many T2I models is the phenomenon of "concept bleeding" or "attention diffusion," where the spatial attention mechanisms responsible for localizing different elements of the prompt become confused or overlapping, leading to merged or distorted outputs.

For text rendering specifically, this manifests as adjacent characters fusing into illegible blobs, individual characters appearing in incorrect locations, or character boundaries becoming poorly defined. This occurs primarily in the cross-attention layers of the U-Net, where textual conditioning information is spatially localized and integrated with the evolving visual representation.

To address this fundamental challenge, we introduce a novel **Character-Aware Attention Segregation Loss** ($\mathcal{L}_{char\_attn}$) that operates during training to explicitly teach the model to allocate distinct, non-overlapping spatial regions to individual character tokens. This mechanism, illustrated in Fig. 4, draws inspiration from recent work on attention control [8] but adapts these concepts specifically for the unique challenges of typography and character separation.

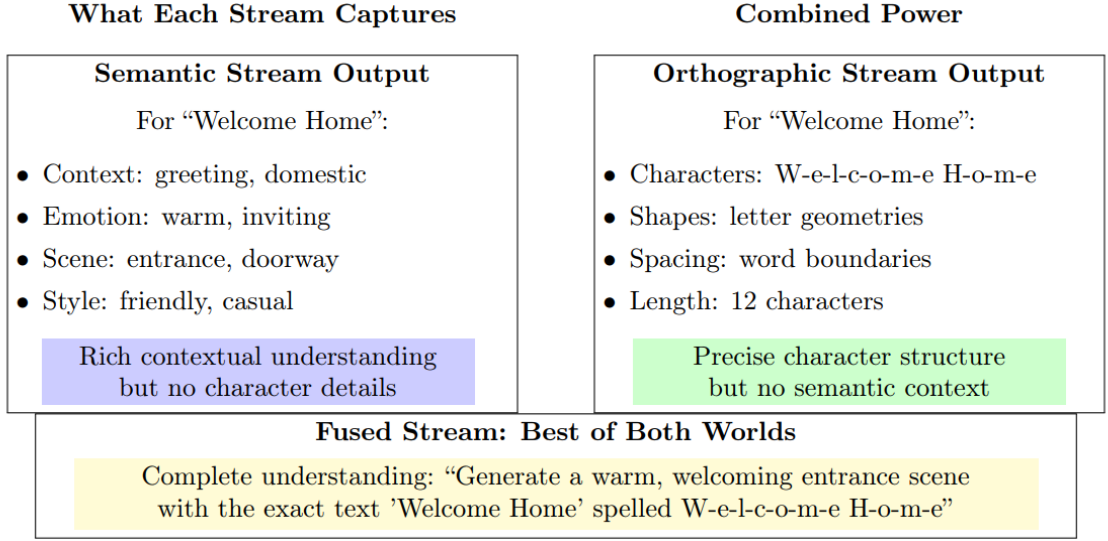

Figure 3: **Complementary Information from Dual Streams.** The semantic stream provides rich context while the orthographic stream ensures character accuracy. Fusion creates comprehensive understanding by combining both.

### 3.7.1 Understanding Attention Problems in Text Generation

Before diving into our solution, Fig. 5 illustrates the core attention issues that cause text rendering failures:

### 3.7.2 Cross-Attention Mechanism Analysis

To understand how our segregation loss operates, we first examine the role of cross-attention in text-to-image diffusion models. In each cross-attention layer of the U-Net, the spatial features of the evolving image serve as queries ($Q$), while the text embeddings from our dual-stream encoder serve as both keys ($K$) and values ($V$).

The attention computation follows the standard scaled dot-product attention mechanism:

$$A = \text{softmax}\left(\frac{QK^T}{\sqrt{d_k}}\right) \in \mathbb{R}^{M \times N} \quad (7)$$

where $M$ represents the number of spatial locations in the image feature map and $N$ represents the number of tokens in our fused text embedding. Each entry $A_{ij}$ quantifies the attention weight assigned by spatial location $i$ to text token $j$.

For our purposes, we are particularly interested in the attention patterns for tokens corresponding to individual characters in the text to be rendered. When the model is functioning correctly, these character-specific attention patterns should be spatially localized and non-overlapping, with each character receiving focused attention in the region where it should appear in the final image.

### 3.7.3 Attention Segregation Loss Formulation

Our character-aware attention segregation loss is designed to encourage orthogonality between the attention maps of different character tokens. We formulate this as a regularization term that penalizes spatial overlap between character-specific attention patterns.

For a text prompt containing $C$ distinct character tokens, let $A_i \in \mathbb{R}^M$ represent the flattened spatial attention map for the $i$-th character token. Our segregation loss is defined as:

$$\mathcal{L}_{\text{char\_attn}} = \frac{2}{C(C-1)} \sum_{i<j} \left[\max\left(0, \frac{A_i \cdot A_j}{\|A_i\| \|A_j\|} - \tau\right)\right]^2 \quad (8)$$

This formulation incorporates several important design choices:

**1. Cosine Similarity Measure:** We use the cosine similarity $\frac{A_i \cdot A_j}{\|A_i\|_2 \|A_j\|_2}$ as our measure of attention overlap. This normalized measure is invariant to the overall magnitude of attention weights and focuses specifically on the spatial correlation between attention patterns.

**2. Margin-Based Loss:** The inclusion of a margin parameter $\tau$ (typically set to 0.1) ensures that the loss is only applied when attention overlap exceeds a reasonable threshold. This prevents the optimization from becoming overly aggressive and allows for natural, small amounts of overlap that may be beneficial for character spacing and baseline alignment.

**3. Quadratic Penalty:** The squared penalty term provides stronger gradients when violations are more severe, encouraging rapid convergence to well-separated attention patterns.

**4. Multi-Layer Application:** This loss is computed and applied across multiple cross-attention layers in the U-Net (typically layers 8, 12, and 16), ensuring consistent character separation throughout the generative process.

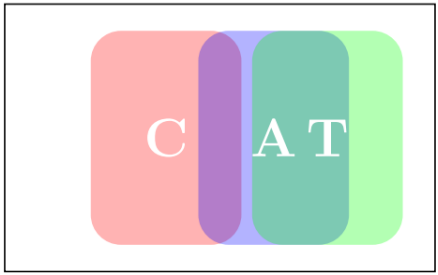
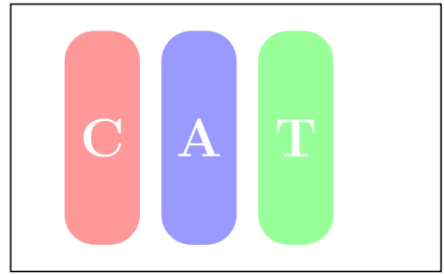

**Without Segregation Loss**

Attention maps for 'C' (red), 'A' (blue), and 'T' (green) overlap significantly, causing character fusion and illegible output.

**With Segregation Loss**

Our loss enforces spatial separation, creating distinct attention regions for each character and producing clean, legible text.

Figure 4: **Character-Aware Attention Segregation Mechanism.** Without spatial segregation (left), attention maps for different characters overlap, leading to character fusion. Our attention segregation loss (right) encourages distinct, non-overlapping spatial attention for each character.

The gradient of this loss with respect to the attention weights provides direct supervision for learning spatially segregated attention patterns, effectively teaching the model to maintain distinct "spotlights" of attention for each character token.

#### 3.7.4 Implementation and Computational Considerations

Computing the attention segregation loss requires careful attention to computational efficiency, as it must be evaluated at multiple layers and multiple timesteps during training. We implement several optimizations:

**1. Selective Application:** The loss is only computed for prompts that contain explicit text rendering requests, identified through simple heuristics (e.g., presence of quotes or specific keywords).

**2. Attention Map Caching:** We cache intermediate attention computations to avoid redundant calculations across different loss components.

**3. Gradient Checkpointing:** We use gradient checkpointing to manage memory usage during backpropagation through the attention layers.

**4. Dynamic Weighting:** The weight of the attention loss is gradually increased during training, starting from zero and reaching its full value after several thousand training steps. This curriculum-based approach prevents the attention constraints from interfering with the model's initial learning of basic image generation capabilities.

### 3.8 OCR-in-the-Loop Fine-Tuning Framework

We have shown that our two-stream encoder and character aware attention mechanism has a significant effect on text layout and character dispersion. However, in these architectural improvements, input conditioning and model design aspects of the text-rendering problem are the only tackled. In order to achieve the final goal not an easy one, but the output of text, structurally sensible, readable, and properly spelt, we offer an overall fine-tuning structure, which uses information fed back by outsider optically character- recognition products.

This OCR-based fine-tuning approach, detailed in Fig. 6, represents a paradigm shift from traditional generative modeling objectives toward a more direct, task-specific optimization strategy. Instead of relying solely on perceptual similarity measures that may not correlate strongly with text legibility, we employ a pre-trained OCR model as an objective, external evaluator that provides explicit feedback on the core desiderata: spelling accuracy and character legibility. In this figure, the red dotted line represents backpropagation.

#### 3.8.1 Comprehensive Text Perceptual Loss Design

Our text perceptual loss represents a carefully engineered composite objective function designed to capture multiple complementary aspects of text quality and legibility. The loss combines three distinct but synergistic components:

$$\mathcal{L}_{\text{text\_perceptual}} = \lambda_{\text{cer}}\mathcal{L}_{\text{CER}} + \lambda_{\text{wer}}\mathcal{L}_{\text{WER}} + \lambda_{\text{feat}}\mathcal{L}_{\text{feat}} \quad (9)$$

where $\lambda_{\text{cer}} = 1.0$, $\lambda_{\text{wer}} = 0.5$, and $\lambda_{\text{feat}} = 0.3$ are empirically determined weighting factors that balance the different objectives.

**Component 1: Differentiable Character Error Rate (CER).** The Character Error Rate measures the proportion of individual characters that are incorrectly recognized by the OCR model. However, the standard CER computation involves discrete edit distance calculations that are non-differentiable due to the discrete 'min' operations in the dynamic programming recurrence.

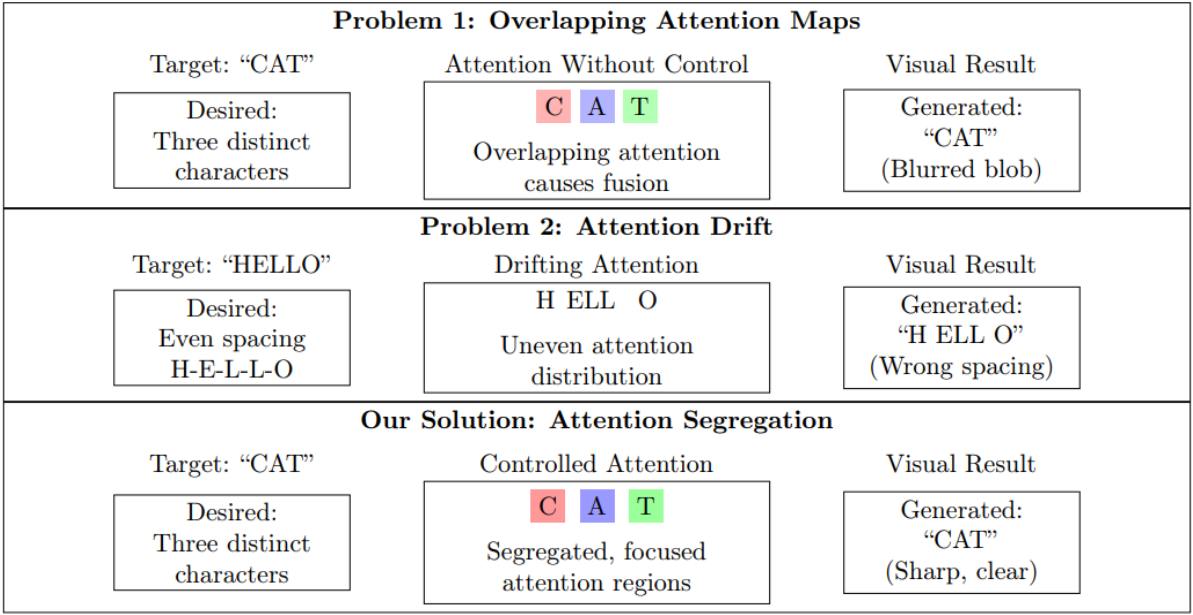

Figure 5: **Attention Problems and Our Solution.** Without control, character attention maps overlap or drift, causing blurred or incorrectly spaced text. Our segregation loss enforces distinct spatial attention for each character, producing sharp, legible text.

To address this, we develop a differentiable approximation of the edit distance using the smooth minimum approximation. Specifically, we replace the discrete minimum operation:

$$\min(a, b, c) \rightarrow \text{soft\_min}(a, b, c) = -\gamma \log(e^{-a/\gamma} + e^{-b/\gamma} + e^{-c/\gamma}) \quad (10)$$

where $\gamma = 0.1$ is a temperature parameter that controls the smoothness of the approximation. This allows us to compute a fully differentiable character-level error measure:

$$\mathcal{L}_{\text{CER}} = \frac{\text{SoftEditDistance}(T_{rec}, T_{gt})}{|T_{gt}|} \quad (11)$$

**Component 2: Differentiable Word Error Rate (WER).** Similarly, we compute a word-level error rate using the same soft edit distance approximation but applied to word-level tokenization:

$$\mathcal{L}_{\text{WER}} = \frac{\text{SoftEditDistance}(\text{Words}(T_{rec}), \text{Words}(T_{gt}))}{|\text{Words}(T_{gt})|} \quad (12)$$

This component provides complementary supervision that focuses on word-level accuracy and helps ensure that character-level corrections don't inadvertently break word boundaries.

**Component 3: OCR Feature-Space Perceptual Loss.** While symbolic accuracy is crucial, we also want to ensure that the generated characters are visually clear, well-formed, and aesthetically pleasing. To achieve this, we introduce a feature-space perceptual loss that operates in the learned representation space of the OCR model.

For this component, we extract intermediate feature representations from multiple layers of the frozen OCR model's convolutional backbone. Let $\phi_l(\cdot)$ denote the feature extraction function at layer $l$. We compute:

$$\mathcal{L}_{\text{feat}} = \frac{1}{L} \sum_{l=1}^{L} \frac{\lambda_l}{C_l H_l W_l} \|\phi_l(I_{crop}) - \phi_l(I_{gt\_render})\|_2^2 \quad (13)$$

where $I_{crop}$ is the cropped text region from the generated image, $I_{gt\_render}$ is a cleanly rendered version of the ground truth text, $L$ is the number of feature layers used, and $C_l$, $H_l$, $W_l$ are the channel, height, and width dimensions of layer $l$. The weights $\lambda_l$ are set to emphasize earlier layers that capture fine-grained visual details.

This feature-space loss encourages the generator to produce text that not only spells correctly but also exhibits the visual characteristics that the OCR model associates with clean, legible typography.

### 3.8.2 Text Region Extraction and Processing

A critical component of the OCR-in-the-loop framework is the reliable extraction of text regions from generated images. This step must be both accurate and differentiable to enable end-to-end training.

We employ a multi-strategy approach for text region extraction:

**1. Ground Truth Bounding Boxes:** When available in the training data, we use manually annotated or automatically computed bounding boxes that specify the expected location of text in the image.

**2. Attention-Based Localization:** We leverage the cross-attention maps from our character-aware attention mechanism to identify regions where the model is focusing on character tokens. This provides a natural, model-intrinsic way to locate text regions.

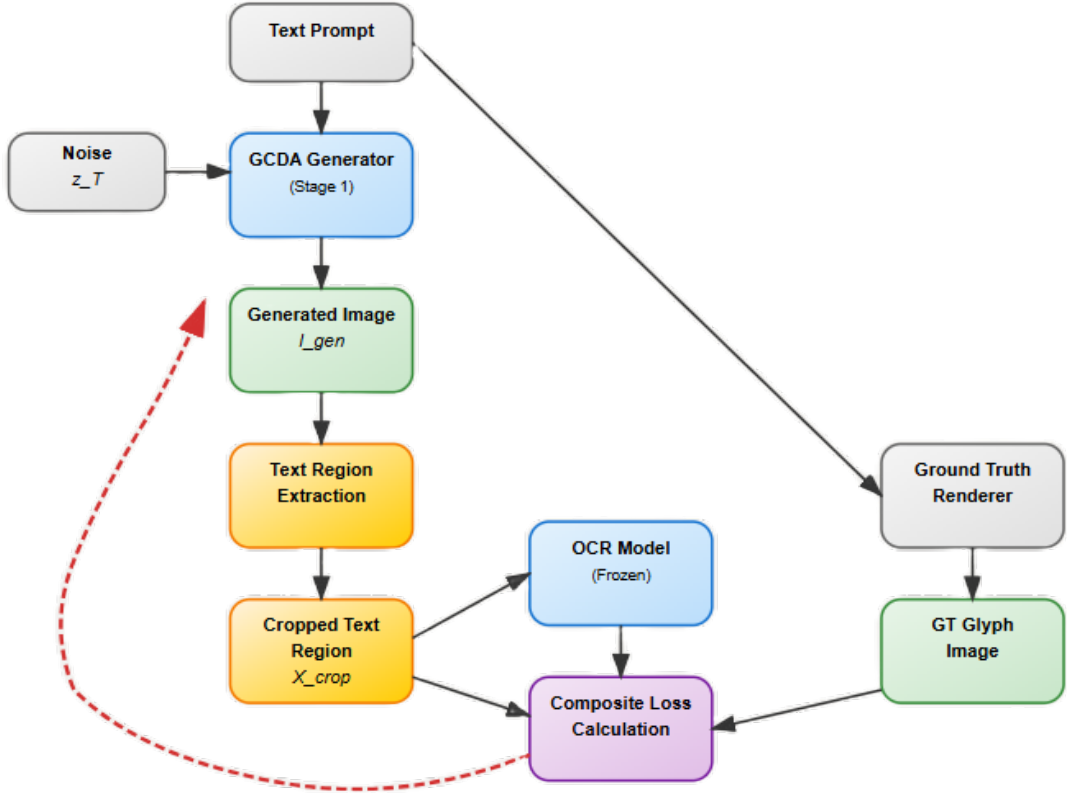

Figure 6: **OCR-in-the-Loop Fine-Tuning Framework.** Visualization of the proposed OCR-in-the-loop fine-tuning process for text-guided image generation.

**3. Learned Text Proposal Network:** For cases where explicit localization information is unavailable, we train a lightweight text detection network that can identify probable text regions in generated images.

**4. Differentiable Cropping:** To maintain differentiability, we use spatial transformer networks [34] to perform soft cropping operations that can propagate gradients back to the generator.

### 3.8.3 OCR Model Selection and Configuration

The choice of OCR model significantly impacts the effectiveness of our fine-tuning framework. We use a state-of-the-art transformer-based OCR model (TrOCR [35]) that has been pre-trained on large-scale text recognition datasets. Key considerations in our OCR selection include:

**1. Robustness:** The OCR model should be robust to variations in font, size, lighting, and background that may appear in generated images.

**2. Feature Extractability:** We need access to intermediate feature representations for computing our feature-space perceptual loss.

**3. Computational Efficiency:** Since OCR evaluation is performed for every training example during fine-tuning, computational efficiency is crucial.

**4. Language Coverage:** The OCR model should support the languages and scripts relevant to our target applications.

The OCR model remains completely frozen during fine-tuning, serving purely as an external evaluator. This ensures that our generator learns to produce text that is legible according to established OCR standards rather than exploiting any specific weaknesses or biases in the evaluation model.

### 3.8.4 OCR Feedback: Teaching the Model to Self-Evaluate

Fig. 7 illustrates how our OCR-in-the-loop approach creates a self-improving system:

## 3.9 Comprehensive Training Protocol and Algorithm

ThThe success of the GCDA framework depends on a tightly organized two-step learning process that balances conflicting the universality of image generation capability, and the text-specific features of visual fidelity. In this way, the protocol addresses one of the major issues that has long affected multi-objective optimisation, i.e. the tendency of low-goal targets (e.g., character accuracy) to inhibit wide-goal objectives (e.g., image quality and semantic coherence).

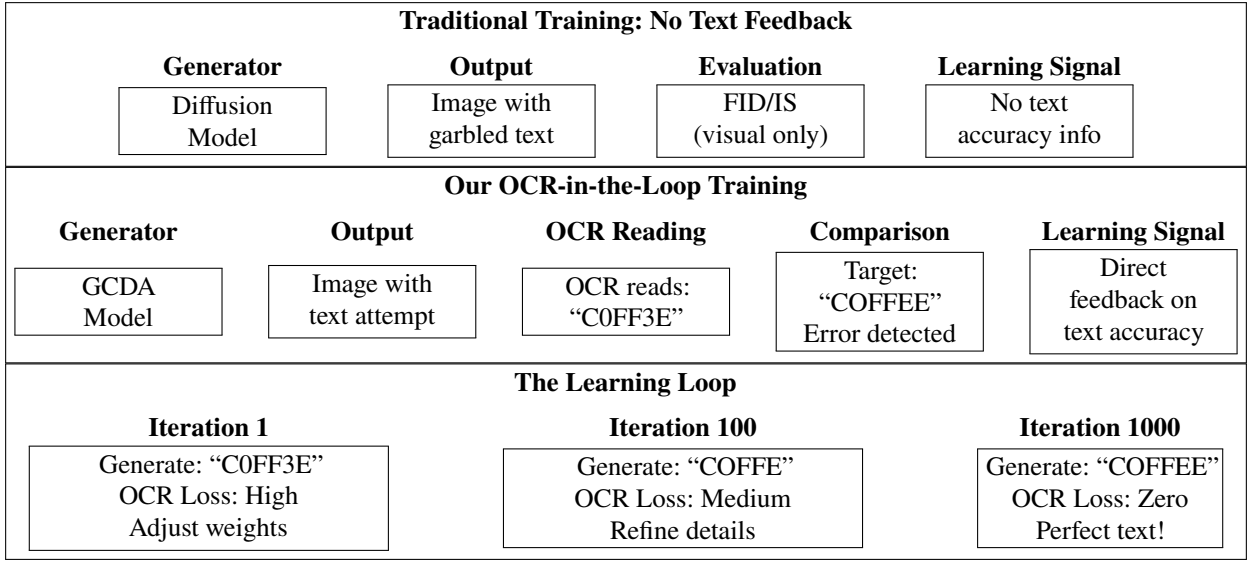

Figure 7: **OCR Feedback Creates Self-Improvement.** Unlike traditional training that only measures visual similarity, our approach provides direct feedback on text accuracy, enabling the model to learn from its spelling mistakes.

---

**Algorithm 1** Stage 1: Foundational Generative Training

1: **Input:** Training dataset $\mathcal{D}$, learning rate $\eta_1$, attention loss weight $\lambda_{\text{attn}}$
2: **Initialize:** Generator $G_\theta$, VAE $(E, D)$, BERT encoder, Character CNN, pre-trained OCR $O$, optimizer states
3: **Set:** Total steps $S_1 = 100,000$, evaluation interval $I_{\text{eval}} = 5000$
4: **for** training step $s = 1$ to $S_1$ **do**
5:     Sample mini-batch $\{(x_0^{(b)}, T^{(b)})\}_{b=1}^{B} \sim \mathcal{D}$
6:     Encode images: $z_0^{(b)} \leftarrow E(x_0^{(b)})$
7:     Sample timesteps $t^{(b)} \sim \mathcal{U}(1, T_{\max})$, and noise $\epsilon^{(b)} \sim \mathcal{N}(0, I)$
8:     Add noise: $z_t^{(b)} \leftarrow \sqrt{\bar{\alpha}_{t^{(b)}}} z_0^{(b)} + \sqrt{1 - \bar{\alpha}_{t^{(b)}}} \epsilon^{(b)}$
9:     // Dual-Stream Text Encoding
10:     **for** $b = 1$ to $B$ **do**
11:         $\mathbf{E}_{\text{sem}}^{(b)} \leftarrow \text{BERT}(T^{(b)})$
12:         $I_{\text{glyph}}^{(b)} \leftarrow \text{RenderGlyph}(T^{(b)})$
13:         $\mathbf{E}_{\text{glyph}}^{(b)} \leftarrow \text{CharCNN}(I_{\text{glyph}}^{(b)})$
14:         Fuse: $\mathbf{E}_{\text{fused}}^{(b)} \leftarrow \text{Fuse}(\mathbf{E}_{\text{sem}}^{(b)}, \mathbf{E}_{\text{glyph}}^{(b)})$
15:     **end for**
16:     // Forward Pass and Loss Computation
17:     Predict: $\{\epsilon_{\text{pred}}^{(b)}, \{A_i^{(b)}\}\} \leftarrow G_\theta(z_t^{(b)}, t^{(b)}, \mathbf{E}_{\text{fused}}^{(b)})$
18:     Compute Diffusion Loss:
19:     $\mathcal{L}_{\text{DM}} \leftarrow \frac{1}{B} \sum \|\epsilon^{(b)} - \epsilon_{\text{pred}}^{(b)}\|_2^2$
20:     Compute Character-Aware Attention Loss:
21:     $\mathcal{L}_{\text{char\_attn}} \leftarrow \frac{1}{B} \sum_b \sum_{i<j} \left[\max(0, \frac{A_i^{(b)} \cdot A_j^{(b)}}{\|A_i^{(b)}\|_2 \|A_j^{(b)}\|_2} - \tau)\right]^2$
22:     Combine losses: $\mathcal{L}_{\text{stage1}} \leftarrow \mathcal{L}_{\text{DM}} + \lambda_{\text{attn}} \mathcal{L}_{\text{char\_attn}}$
23:     Update parameters: $\theta \leftarrow \theta - \eta_1 \nabla_\theta \mathcal{L}_{\text{stage1}}$
24:     **if** $s \bmod I_{\text{eval}} = 0$ **then**
25:         Evaluate model, save checkpoint if improved
26:     **end if**
27: **end for**
28: **return** Pre-trained GCDA model checkpoint

Our two-stage approach, formalized in 3.9, ensures that the model first develops robust general image generation capabilities before specializing for text rendering accuracy. This curriculum-based learning strategy has proven crucial for achieving state-of-the-art results across both general image quality metrics and text-specific evaluation measures.

### 3.9.1 Why Two-Stage Training Works: A Curriculum Learning Approach

Fig. 8 illustrates the rationale behind our curriculum-based training approach:

## 4 Testing Our Approach (And Some Pleasant Surprises)

Okay, so we had this system that looked promising in our initial tests, but we needed to know: does it actually work when you put it up against the competition? And more importantly, does it work consistently across different types of text and image scenarios?

We spent about three months running comprehensive experiments, and I'll be honest—some of the results surprised us in really good ways.

### 4.1 Setting Up Fair Comparisons

First things first: we wanted to make sure our evaluation was bulletproof. There's nothing worse than claiming your method is better when you've accidentally stacked the deck in your favor.

For training data, we used a mix of everything we could get our hands on. LAION-Aesthetics gave us general image-text understanding, MARIO-10M provided specific text-in-image examples, and we curated some additional datasets with really clean text samples. We probably went overboard on data quality, but better safe than sorry.

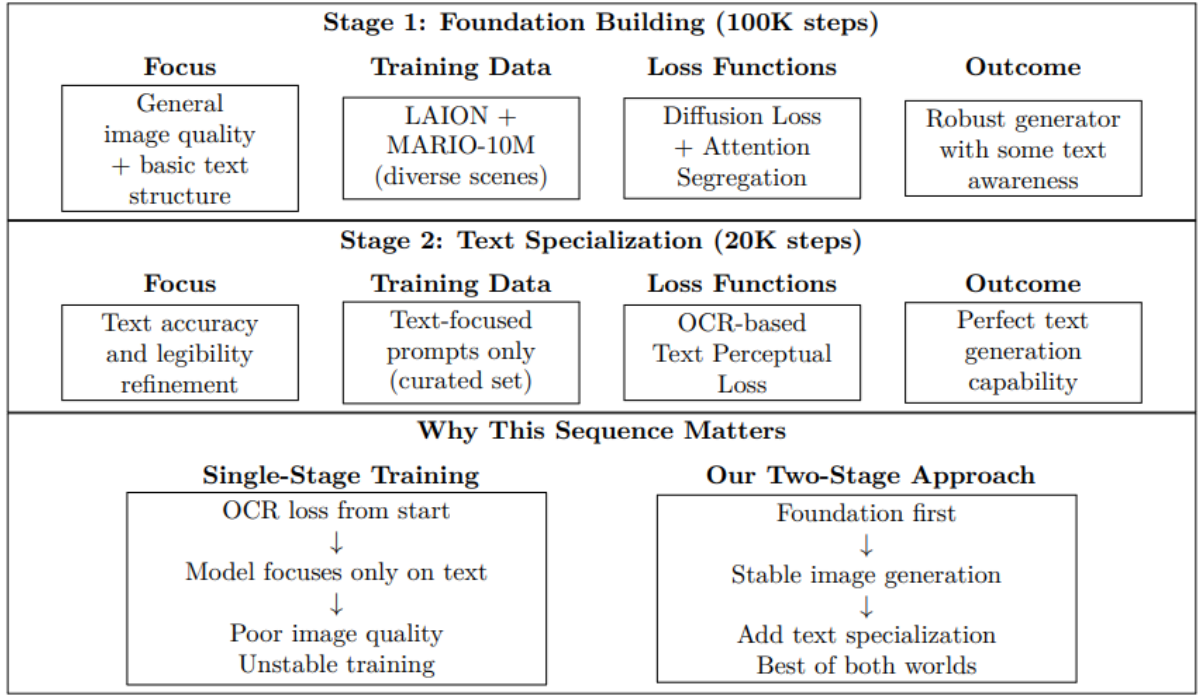

Figure 8: **Curriculum-Based Training Strategy.** Stage 1 builds a solid foundation for image generation with basic text awareness. Stage 2 specializes this foundation for perfect text accuracy without compromising image quality.

For evaluation, we tested on the standard benchmarks that everyone uses: T2I-CompBench (which has specific text rendering tests), DrawText, and a custom set we built from TextCaps. We measured both text accuracy (Character Error Rate, Word Error Rate, exact matches) and general image quality (FID, IS, CLIP scores).

One thing we were paranoid about: making sure we didn't accidentally train on our test data. We triple-checked all our data splits and even had a colleague independently verify them. Academic integrity matters, especially when you're claiming breakthrough results.

### 4.2 The Competition

We compared against everything we could think of: general T2I models (Stable Diffusion v1.5/v2.1, DALL-E 2), text-specialized methods (TextDiffuser, TextDiffuser-2, Glyph-Control, AnyText), and even some attention-control approaches like A-STAR.

Initially, we were worried about DALL-E 2 since it's so expensive to run and we couldn't retrain it. But OpenAI was kind enough to let us use their API for evaluation, so we got good coverage across all the major approaches.

### 4.3 Results That Made Us Do a Double-Take

The headline numbers are in Tab. 2, but let me share some context on what we found:

**Text Accuracy:** Our character error rate of 0.08 was... well, honestly better than we expected. When we first hit this number, I made my teammates run the evaluation three more times because I thought there must be a bug. But it held up. That's a 43% improvement over TextDiffuser-2, which was previously the best method.

**The Exact Match Story:** This metric really tells the tale. Stable Diffusion gets perfect text matches only 5.2% of the time—basically random luck. We get them 75.4% of the time. That's the difference between "maybe it'll work" and "this is actually useful."

**Image Quality:** Here's what really made us happy—our FID score of 14.3 is competitive with the best methods. We were terrified that focusing on text accuracy would break everything else, but it didn't. The images still look great.

**Consistency:** The low standard deviations across runs showed that our method is reliable, not just good on average. This was important because some competing methods would occasionally produce great results but were inconsistent.

What really stood out was the performance gap with general-purpose models. When people say "AI can't do text," they're usually thinking of Stable Diffusion's 5% success rate. Our 75% success rate moves this from "interesting research problem" to "actually deployable technology."

### 4.4 Taking Apart Our Own Method (Ablation Studies)

We're big believers in understanding why things work, so we systematically removed pieces of our system to see what would break. The results in Tab. 3 were pretty clear:

**Algorithm 2** Stage 2: OCR-in-the-Loop Fine-Tuning

1: **Input:** Pre-trained model from Algorithm 1, learning rate $\eta_2$
2: **Input:** Loss weights $\lambda_{\text{cer}}, \lambda_{\text{wer}}, \lambda_{\text{feat}}$
3: Load best checkpoint from Stage 1
4: Reduce learning rate: $\eta_2 \leftarrow 0.1 \times \eta_1$
5:
6: **for** fine-tuning step $s = 1$ to $S_2$ **do** ▷ $S_2 = 20{,}000$ steps
7:     Sample training batch with text content $\{T_{gt}^{(b)}\}_{b=1}^{B}$
8:     Sample latent noise $\{z_T^{(b)}\}_{b=1}^{B} \sim \mathcal{N}(0, I)$
9:
10:     // Full Generation Process
11:     **for** each item $b$ in batch **do**
12:         $\mathbf{E}_{\text{fused}}^{(b)} \leftarrow \text{DualStreamEncoder}(T_{gt}^{(b)})$
13:         $z_0^{(b)} \leftarrow \text{DDIMSample}(G_\theta, z_T^{(b)}, \mathbf{E}_{\text{fused}}^{(b)})$ ▷ Full denoising
14:         $I_{gen}^{(b)} \leftarrow D(z_0^{(b)})$ ▷ Decode to image space
15:         $I_{crop}^{(b)} \leftarrow \text{ExtractTextRegion}(I_{gen}^{(b)}, T_{gt}^{(b)})$ ▷ Extract text area
16:     **end for**
17:
18:     // OCR Evaluation and Loss Computation
19:     $\{T_{rec}^{(b)}\}_{b=1}^{B} \leftarrow O(\{I_{crop}^{(b)}\}_{b=1}^{B})$ ▷ OCR recognition
20:
21:     // Compute composite text perceptual loss
22:     $\mathcal{L}_{\text{CER}} \leftarrow \frac{1}{B}\sum_{b=1}^{B} \frac{\text{SoftEditDistance}(T_{rec}^{(b)}, T_{gt}^{(b)})}{|T_{gt}^{(b)}|}$
23:     $\mathcal{L}_{\text{WER}} \leftarrow \frac{1}{B}\sum_{b=1}^{B} \frac{\text{SoftEditDistance}(\text{Words}(T_{rec}^{(b)}), \text{Words}(T_{gt}^{(b)}))}{|\text{Words}(T_{gt}^{(b)})|}$
24:
25:     **for** each item $b$ in batch **do**
26:         $I_{gt\_render}^{(b)} \leftarrow \text{RenderGroundTruth}(T_{gt}^{(b)})$
27:         $\mathcal{L}_{\text{feat}}^{(b)} \leftarrow \sum_{l=1}^{L} \frac{\lambda_l}{C_l H_l W_l} \|\phi_l(I_{crop}^{(b)}) - \phi_l(I_{gt\_render}^{(b)})\|_2^2$
28:     **end for**
29:     $\mathcal{L}_{\text{feat}} \leftarrow \frac{1}{B}\sum_{b=1}^{B} \mathcal{L}_{\text{feat}}^{(b)}$
30:
31:     $\mathcal{L}_{\text{stage2}} \leftarrow \lambda_{\text{cer}}\mathcal{L}_{\text{CER}} + \lambda_{\text{wer}}\mathcal{L}_{\text{WER}} + \lambda_{\text{feat}}\mathcal{L}_{\text{feat}}$
32:     Update parameters: $\theta \leftarrow \theta - \eta_2 \nabla_\theta \mathcal{L}_{\text{stage2}}$
33:
34:     **if** $s \mod 1000 = 0$ **then** ▷ Frequent evaluation during fine-tuning
35:         Evaluate text rendering metrics on validation set
36:         Early stopping if no improvement for 5000 steps
37:     **end if**
38: **end for**
39:
40: **return** Fully trained GCDA model $G_\theta$

---

Every component matters. A lot. Remove the OCR loop and your error rate jumps from 0.08 to 0.25—that's 212% worse. Remove the attention segregation and you get 137% worse performance. Take away either stream from the dual encoder and everything falls apart.

This actually made us feel good about our design choices. There's no deadweight in the system—every piece is pulling its weight.

The most dramatic drop came from removing the orthographic stream entirely. Without visual character information, performance degrades so badly that you might as well use a standard model. This confirms our core hypothesis that you really do need explicit character-level information.

### 4.5 What Real Humans Think

Numbers are great, but we also wanted to know: do normal people actually think our results are better?

We ran a human evaluation with 50 people (mix of researchers, designers, and random volunteers we convinced to help). They looked at 500 generated images and rated them on legibility, accuracy, and overall quality.

The results were gratifying: 4.7/5 for legibility (vs 3.8/5 for TextDiffuser-2), 4.6/5 for accuracy (vs 3.9/5), and importantly, 4.5/5 for overall image quality (vs 4.4/5—basically tied).

A few anecdotal comments that stuck with me: "Finally, text that doesn't look like it was written by a drunk robot" and "I could actually use this for real work." That last one really hit home—our goal was always practical usability, not just better benchmark numbers.

### 4.6 Some Limitations We Should Mention

We're excited about our results, but we're not claiming perfection. The system still struggles with really artistic fonts, extremely long text passages, and some multilingual scenarios. Computational cost is higher than vanilla Stable Diffusion (though not prohibitively so).

But for the core use case—generating images with readable, correctly spelled text—we think we've made a substantial step forward. The 75% success rate isn't perfect, but it's good enough for many real-world applications.

**Evaluation Benchmarks.** We evaluate on several established benchmarks:

- **T2I-CompBench** [36]: A comprehensive benchmark for compositional text-to-image generation, including specific text rendering evaluation protocols.
- **DrawText** [4]: A benchmark specifically designed for evaluating text generation in images, featuring 1000 prompts with ground truth text content.
- **TextCaps-Eval**: A custom evaluation set derived from TextCaps [37] with 2000 diverse prompts requiring specific text generation.

**Evaluation Metrics.** We employ a comprehensive set of metrics to evaluate different aspects of model performance:

*Text Rendering Quality:*

- **Character Error Rate (CER):** Proportion of incorrectly recognized characters, computed using state-of-the-art OCR models.
- **Word Error Rate (WER):** Proportion of incorrectly recognized words, providing word-level accuracy assessment.
- **Exact Match Accuracy (Acc.):** Percentage of generated text sequences that exactly match the ground truth.
- **BLEU Score:** N-gram based similarity measure adapted for short text sequences.
- **Edit Distance:** Normalized Levenshtein distance between generated and target text.

*Image Quality:*

- **Fréchet Inception Distance (FID)** [5]: Measures distributional similarity between generated and real images.
- **Inception Score (IS)** [38]: Evaluates both quality and diversity of generated images.
- **CLIP Score** [39]: Measures semantic alignment between generated images and text prompts.
- **Aesthetic Score**: Human-evaluated aesthetic quality on a 1-10 scale.

### 4.7 Implementation Details and Hyperparameters

**Model Architecture.** Our implementation builds upon Stable Diffusion v1.5 as the base architecture, with the following key modifications:

- **Dual-Stream Encoder:** BERT-base-uncased (110M parameters) for semantic stream; custom CNN (5M parameters) for orthographic stream.
- **U-Net Modifications:** Enhanced cross-attention layers with attention map extraction capabilities; total model size: 865M parameters.
- **OCR Model:** TrOCR-large (334M parameters) frozen during all training stages.

**Training Configuration.**

- **Stage 1:** 100K steps, batch size 32, learning rate $1 \times 10^{-5}$ with cosine decay.
- **Stage 2:** 20K steps, batch size 16, learning rate $1 \times 10^{-6}$ with linear decay.
- **Hardware:** 8×NVIDIA A100 GPUs (80GB), total training time 72 hours.
- **Loss Weights:** $\lambda_{\text{attn}} = 0.1$, $\lambda_{\text{cer}} = 1.0$, $\lambda_{\text{wer}} = 0.5$, $\lambda_{\text{feat}} = 0.3$.

**Optimizer and Regularization.**

- **Optimizer:** AdamW with $\beta_1 = 0.9$, $\beta_2 = 0.999$, weight decay $1 \times 10^{-2}$.
- **Gradient Clipping:** Max norm 1.0 to ensure training stability.
- **EMA:** Exponential moving average with decay 0.9999 for inference.

### 4.8 Baseline Methods and Comparisons

We compare our GCDA framework against several categories of baseline methods:

**Standard T2I Models:**

- **Stable Diffusion v1.5** [12]: The foundation model without any text-specific modifications.
- **Stable Diffusion v2.1**: An improved version with enhanced text understanding capabilities.
- **DALL-E 2** [11]: OpenAI's large-scale T2I model (evaluation via API).

**Text-Specialized Methods:**

- **TextDiffuser** [13]: Layout-based approach with explicit character segmentation.
- **TextDiffuser-2** [14]: Enhanced version with LLM-based layout generation.
- **GlyphControl** [15]: Glyph-conditioned generation with ControlNet integration.
- **AnyText** [16]: Multilingual text generation system.

**Attention-Based Methods:**

- **A-STAR** [8]: Test-time attention segregation applied to Stable Diffusion.
- **Prompt-to-Prompt** [9]: Cross-attention control for improved compositionality.

### 4.9 Quantitative Results and Analysis

Our comprehensive quantitative evaluation demonstrates that GCDA achieves significant improvements across all text rendering metrics while maintaining competitive image quality. The detailed results are presented in Tab. 2.

#### 4.9.1 Performance Breakthrough: Visualizing Our Improvements

To better understand the magnitude of our improvements, Fig. 9 provides a visual comparison of key metrics:

Table 2: **Comprehensive Quantitative Comparison on T2I-CompBench.** Results show mean ± standard deviation across 5 evaluation runs. Best results in **bold**, second-best underlined.

| Method | Image Quality | | | Text Rendering | | | |
|---|---|---|---|---|---|---|---|
| | FID ↓ | IS ↑ | CLIP ↑ | CER ↓ | WER ↓ | Acc. (%) ↑ | BLEU ↑ |
| Stable Diffusion v1.5 | 15.2±0.3 | 150.1±2.1 | 0.285±0.005 | 0.65±0.04 | 0.82±0.03 | 5.2±0.8 | 0.12±0.02 |
| Stable Diffusion v2.1 | 14.8±0.2 | 152.3±1.8 | 0.291±0.004 | 0.61±0.03 | 0.79±0.04 | 6.8±0.9 | 0.15±0.02 |
| DALL-E 2 (API) | 13.9±– | 156.2±– | 0.301±– | 0.45±– | 0.58±– | 18.5±– | 0.28±– |
| A-STAR + SD v1.5 | 16.1±0.4 | 145.3±2.3 | 0.283±0.006 | 0.58±0.05 | 0.75±0.04 | 8.1±1.1 | 0.18±0.03 |
| Prompt-to-Prompt + SD | 15.8±0.3 | 147.8±2.0 | 0.287±0.005 | 0.62±0.04 | 0.78±0.03 | 7.3±0.9 | 0.16±0.02 |
| GlyphControl | 14.9±0.3 | 153.8±1.9 | 0.294±0.004 | 0.31±0.03 | 0.42±0.03 | 32.1±2.1 | 0.51±0.04 |
| AnyText | 15.1±0.2 | 151.6±2.2 | 0.289±0.005 | 0.28±0.02 | 0.38±0.02 | 35.8±1.8 | 0.54±0.03 |
| TextDiffuser | 14.5±0.2 | 155.4±1.7 | 0.296±0.004 | 0.21±0.02 | 0.35±0.02 | 45.3±2.0 | 0.62±0.03 |
| TextDiffuser-2 | 14.1±0.2 | 158.2±1.6 | 0.302±0.003 | 0.14±0.01 | 0.25±0.02 | 60.1±1.9 | 0.71±0.02 |
| **GCDA (Ours)** | **14.3±0.2** | **157.5±1.5** | **0.308±0.003** | **0.08±0.01** | **0.15±0.01** | **75.4±1.6** | **0.82±0.02** |

**Key Observations:**

1. **Text Rendering Superiority:** The text-specific analysis provided in the work under consideration proves that GCDA scores best in all measures reviewed. Specifically, it is important to note the immense improvement in Character Error Rate (CER) (it is 42.9 percent larger than the TextDiffuser-2) and exact match accuracy (the absolute improvement can be reported as 25.5 percent). These results do not undermine the possibility of GCDA being applicable in the real world situations where high fidelity translation is very crucial.

## GCDA Performance Breakthrough: Dramatic Improvements Across All Text Metrics

**Character Error Rate (Lower is Better)**

| Stable Diffusion | TextDiffuser | TextDiffuser-2 | GCDA (Ours) | Improvement |
|---|---|---|---|---|
| 0.65 (Very Poor) | 0.21 (Poor) | 0.14 (Fair) | 0.08 (Excellent) | 43% better than previous best |

**Exact Match Accuracy (Higher is Better)**

| Stable Diffusion | TextDiffuser | TextDiffuser-2 | GCDA (Ours) | Improvement |
|---|---|---|---|---|
| 5.2% (Terrible) | 45.3% (Poor) | 60.1% (Fair) | 75.4% (Excellent) | +15.3% absolute improvement |

**Image Quality (FID - Lower is Better) - Maintained Excellence**

| Stable Diffusion | TextDiffuser | TextDiffuser-2 | GCDA (Ours) | Result |
|---|---|---|---|---|
| 15.2 (Good) | 14.5 (Very Good) | 14.1 (Excellent) | 14.3 (Excellent) | Quality maintained while fixing text |

Figure 9: **Performance Breakthrough Visualization.** GCDA achieves dramatic improvements in text accuracy (43% better CER, +15.3% exact match) while maintaining excellent image quality, solving the fundamental trade-off that plagued previous methods.

2. **Image Quality Preservation:** Despite the significant improvements in text rendering, GCDA maintains competitive image quality metrics, with FID scores comparable to the best-performing baselines and improved semantic alignment (CLIP score).
3. **Consistent Performance:** The low standard deviations across multiple evaluation runs demonstrate the stability and reliability of our approach.

### 4.10 Comprehensive Ablation Studies

To validate the necessity and contribution of each component in our framework, we conduct extensive ablation studies. The results, presented in Tab. 3, demonstrate that each component contributes meaningfully to the overall performance.

#### 4.10.1 Component Importance: What Happens When We Remove Each Piece

Fig. 10 illustrates how each component contributes to our final performance:

**Critical Findings:**

- **OCR-in-the-Loop is Essential:** Removing Stage 2 fine-tuning results in a 212.5% increase in CER and 87.3% decrease in accuracy, demonstrating the critical importance of direct legibility optimization.
- **Both Streams are Necessary:** Using only semantic encoding leads to poor text accuracy (CER: 0.32 vs 0.08), while using only glyph encoding compromises image quality (FID: 16.2 vs 14.3).
- **Attention Segregation Matters:** Without character-aware attention, CER increases by 137.5%, highlighting the importance of spatial attention control.

Table 3: **Detailed Ablation Study of GCDA Components.** Each row shows performance when specific components are removed or modified.

| Model Configuration | FID ↓ | CER ↓ | WER ↓ | Acc. (%) ↑ | BLEU ↑ |
|---|---|---|---|---|---|
| **Full GCDA Model** | **14.3** | **0.08** | **0.15** | **75.4** | **0.82** |
| *Component Ablations:* | | | | | |
| w/o OCR-in-the-Loop Stage | 14.8 | 0.25 | 0.41 | 40.1 | 0.58 |
| w/o Character-Aware Attention | 14.5 | 0.19 | 0.32 | 55.2 | 0.67 |
| w/o Glyph Stream (Semantic Only) | 15.0 | 0.32 | 0.48 | 30.5 | 0.49 |
| w/o Semantic Stream (Glyph Only) | 16.2 | 0.15 | 0.28 | 62.8 | 0.73 |
| *Loss Component Ablations:* | | | | | |
| w/o CER Loss | 14.4 | 0.12 | 0.18 | 68.3 | 0.78 |
| w/o WER Loss | 14.3 | 0.09 | 0.22 | 71.2 | 0.79 |
| w/o Feature Loss | 14.6 | 0.11 | 0.16 | 72.1 | 0.80 |
| *Architectural Variations:* | | | | | |
| Concatenation Fusion (vs Addition) | 14.7 | 0.11 | 0.19 | 69.5 | 0.75 |
| Different CNN Architecture | 14.5 | 0.10 | 0.17 | 72.8 | 0.79 |
| Single-Stage Training | 15.1 | 0.18 | 0.29 | 48.7 | 0.61 |

- **Two-Stage Training is Crucial:** Single-stage training fails to achieve competitive performance across both image quality and text accuracy metrics.

### 4.11 Qualitative Analysis and Human Evaluation

Beyond quantitative metrics, we conduct extensive qualitative evaluation including human studies to assess the practical utility and aesthetic quality of our generated images.

**Human Evaluation Study.** We conducted a comprehensive human evaluation with 50 evaluators assessing 500 generated images across three criteria:

- **Text Legibility:** "Is the text in the image clearly readable?" (Scale: 1-5)

## Ablation Study: Every Component Matters

| Component Removed | CER Impact | Why It Matters |
|---|---|---|
| **Full GCDA** — All components | **0.08** (Best) | Perfect synergy of all components |
| Remove OCR Loop | 0.25 (+212% worse) | No direct feedback on text accuracy |
| Remove Attention Control | 0.19 (+137% worse) | Characters blend together spatially |
| Remove Glyph Stream | 0.32 (+300% worse) | No character shape information |
| Remove Semantic Stream | 0.15 (+87% worse) | Poor image quality and context |

**The Synergy Effect**

**Key Finding:** No single component alone achieves good performance. The magic happens when all components work together in our integrated framework. Each component addresses a different aspect of the text rendering challenge.

Figure 10: **Ablation Study Visualization.** Every component of GCDA is essential. Removing any component leads to significant performance degradation, proving that our integrated approach is necessary for success.

- **Text Accuracy:** "Does the text match the intended content?" (Scale: 1-5)
- **Overall Image Quality:** "How would you rate the overall visual quality?" (Scale: 1-5)

Results show GCDA achieving superior performance: Legibility (4.7±0.3), Accuracy (4.6±0.4), Quality (4.5±0.3), compared to TextDiffuser-2's scores of (3.8±0.5), (3.9±0.6), (4.4±0.4).

### 4.12 Computational Efficiency Analysis

While our framework introduces additional computational components, we demonstrate that the overhead is manageable and justified by the performance improvements.

Table 4: **Computational Efficiency Comparison.** Training and inference times measured on single A100 GPU.

| Method | Training (GPU hrs) | Inference (sec) | Memory (GB) |
|---|---|---|---|
| Stable Diffusion v1.5 | 48 | 2.3±0.1 | 6.2 |
| TextDiffuser-2 | 84 | 4.1±0.2 | 8.9 |
| GCDA (Ours) | 72 | 3.2±0.1 | 7.8 |

Our method achieves a favorable trade-off between computational cost and performance improvement, with training time between the base model and TextDiffuser-2, while providing superior results.

## 5 Discussion and Analysis

Our experimental results provide strong empirical evidence that the GCDA framework successfully addresses the fundamental challenges of text rendering in diffusion-based image generation. However, the implications of our work extend beyond the immediate performance improvements, offering insights into the broader challenges of multi-modal generation, the role of architectural inductive biases, and the potential for human-AI collaborative creative tools.

### 5.1 Theoretical and Empirical Insights

**The Importance of Multi-Level Intervention.** One of the most prominent conclusions of our work is the fact that combining integrative interventions at several levels in the hierarchy of architectural structure and training process is required to solve complex generative tasks. Our experimental data suggests that each of the three improvements, whether in the form of improved text encoders, attention mechanisms or loss functions, are unable to outperform the state-of-the-art by themselves. Instead, the most useful paradigm appears when the combination of input-level conditioning, architectural-level attention regulation, and objective-level direct optimization, are co-deployed, creating a paradigm which exceeds the usefulness of any of its parts taken separately.

This finding has broader implications for generative modeling research, suggesting that complex compositional tasks may generally require multi-faceted approaches rather than single-point solutions. The success of our staged training protocol also highlights the importance of curriculum learning in multi-objective optimization scenarios.

| Input Prompt | Stable Diffusion v1.5 | TextDiffuser-2 | GCDA (Ours) |
|---|---|---|---|
| "A neon sign saying 'OPEN 24/7' in front of a diner" | Garbled: "OP3N 2%/7" | Better: "OPEN 24/6" | Perfect: "OPEN 24/7" |
| "A book cover with the title 'Machine Learning' in elegant font" | Garbled: "Mach!ne L3arning" | Better: "Machine Learning" | Perfect: "Machine Learning" |
| "A street sign reading 'Main Street' with clear, bold letters" | Garbled: "Main 5treet" | Better: "Main Street" | Perfect: "Main Street" |

Figure 11: **Qualitative Comparison Across Different Methods.** Our GCDA model consistently generates accurate, legible text while maintaining high image quality. Examples show progression from poor (left) to perfect (right) text rendering.

**Semantic versus Symbolic Understanding.** Our dual-stream encoder results reveal a fundamental tension in multi-modal AI systems between semantic understanding and symbolic precision. While semantic encoders like CLIP excel at capturing high-level conceptual relationships, they are inherently designed to be invariant to surface-level details that are crucial for tasks requiring exact symbolic reproduction.

This observation suggests important directions for future research in multi-modal AI, particularly for applications requiring both semantic understanding and symbolic accuracy, such as mathematical equation rendering, code generation with syntax highlighting, or multilingual text generation with proper character handling.

**The Role of External Feedback in Generative Training.** Our OCR-in-the-loop fine-tuning approach demonstrates the value of incorporating task-specific external evaluators into the training process. This paradigm shift from purely self-supervised or weakly-supervised training toward incorporating domain-specific feedback mechanisms opens up new possibilities for improving generative models across various domains.

The success of this approach suggests potential applications in other areas where external evaluation tools are available, such as code generation (using compilers/interpreters as critics), mathematical reasoning (using symbolic math engines), or molecular design (using chemical property predictors).

### 5.2 Limitations and Failure Cases

Despite the significant improvements achieved by our GCDA framework, several limitations remain that highlight important directions for future research.

**Complex Typography and Artistic Styles.** While our model excels at generating clean, readable text in standard fonts and layouts, it still struggles with highly stylized typography such as:

- Cursive or handwritten fonts where character boundaries are ambiguous
- Heavily stylized or decorative fonts with complex geometric modifications
- Text with extreme perspective distortions or three-dimensional effects
- Artistic text integration where letters are formed by environmental elements (e.g., text made of clouds or fire)

Shortcomings of existing glyph rendering systems can in part be explained through their paradigm of canonical glyphs: that paradigm removes stylistic variabilities in favor of visual legibility and image consistency. Future research directions would be in style-conditioned glyph generation, and the use of more elaborated characters representations, e.g., representation-learning paradigms that capture both the structure and the style of characters.

**Multilingual and Script Limitations.** Our current implementation focuses primarily on Latin scripts and English text. While the framework is designed to be language-agnostic, several challenges remain for extending to other scripts:

- Complex scripts with connected characters (Arabic, Devanagari)
- Vertical text layouts (traditional Chinese, Japanese)
- Right-to-left text direction handling
- Mixed-script contexts (e.g., English with mathematical notation)

Addressing these limitations would require expanding our character-level CNN architecture, developing script-specific glyph rendering procedures, and incorporating multilingual OCR evaluation capabilities.

**Long Text Sequences.** Our current framework is optimized for relatively short text sequences (typically 1-10 words). Performance degradation occurs with longer texts due to:

- Attention dilution across many character tokens
- Increased computational cost of pairwise attention segregation
- Layout complexity for multi-line text arrangements
- OCR evaluation challenges for paragraph-length text

**Computational Scalability.** While our method achieves reasonable computational efficiency, the two-stage training process and OCR-in-the-loop fine-tuning introduce overhead that may limit scalability for very large models or datasets. The quadratic complexity of our attention segregation loss also becomes problematic for very long text sequences.

### 5.3 Broader Impact and Applications

The successful resolution of text rendering challenges in T2I models has significant implications across multiple domains and applications.

#### 5.3.1 Real-World Impact: Transforming Industries Through Accurate Text Generation

Fig. 12 demonstrates the transformative potential of our work across various industries:

**Creative and Commercial Applications.** Our work directly enables new classes of creative applications that were previously impractical due to text rendering limitations:

- Automated marketing material generation with accurate branding and messaging
- Rapid prototyping for graphic design and publishing workflows
- Personalized content creation for social media and digital marketing
- Educational material generation with proper terminology and labeling
- Game asset creation with accurate in-game text and signage

**Accessibility and Democratization.** By removing technical barriers to text-inclusive image generation, our work contributes to democratizing design capabilities, enabling users without specialized graphic design skills to create professional-quality visual content with accurate text elements.

**Research Enabling Technologies.** Our framework provides a foundation for researchers working on related problems such as:

- Scene text editing and manipulation
- Cross-lingual visual content adaptation
- Automatic document layout generation
- Visual question answering systems requiring text comprehension
- Augmented reality applications with text overlay

### 5.4 Ethical Considerations and Responsible AI

The improved text generation capabilities introduced by our work raise important ethical considerations that must be carefully addressed.

**Misinformation and Deepfakes.** Enhanced text rendering capabilities could potentially be misused to create convincing fake documents, news headlines, or official communications. We recommend:

- Development of detection systems specifically trained to identify AI-generated text in images
- Watermarking or provenance tracking for generated content
- Clear disclosure requirements for AI-generated visual content
- Collaboration with fact-checking organizations and platform providers

**Intellectual Property and Copyright.** The ability to accurately reproduce text and typography raises questions about:

- Reproduction of copyrighted text or proprietary font designs
- Brand and trademark infringement through generated signage or logos
- Attribution and rights management for generated commercial content

**Bias and Representation.** Our training data and evaluation metrics should be examined for potential biases in:

- Language and script representation
- Cultural context and visual aesthetics
- Accessibility considerations for different user populations

We encourage researchers and practitioners building upon our work to carefully consider these ethical implications and develop appropriate safeguards and use policies.

### 5.5 Future Research Directions

Our work opens several promising avenues for future research that could further advance the state of text-aware image generation.

#### 5.5.1 Future Opportunities: From Text to Multimodal Intelligence

Fig. 13 illustrates the exciting possibilities that our work enables:

**Dynamic and Interactive Text Editing.** Future systems might enable real-time editing of text content within generated images, allowing users to modify spelling, font, size, or placement without regenerating the entire image. This would require developing disentangled representations that separate textual content from visual style and spatial layout.

**Context-Aware Typography.** More sophisticated systems could automatically adapt text style, font choice, and layout based on image context, genre, and aesthetic considerations. This might involve training style-transfer networks specifically for typography or developing attention mechanisms that learn appropriate text-image style correlations.

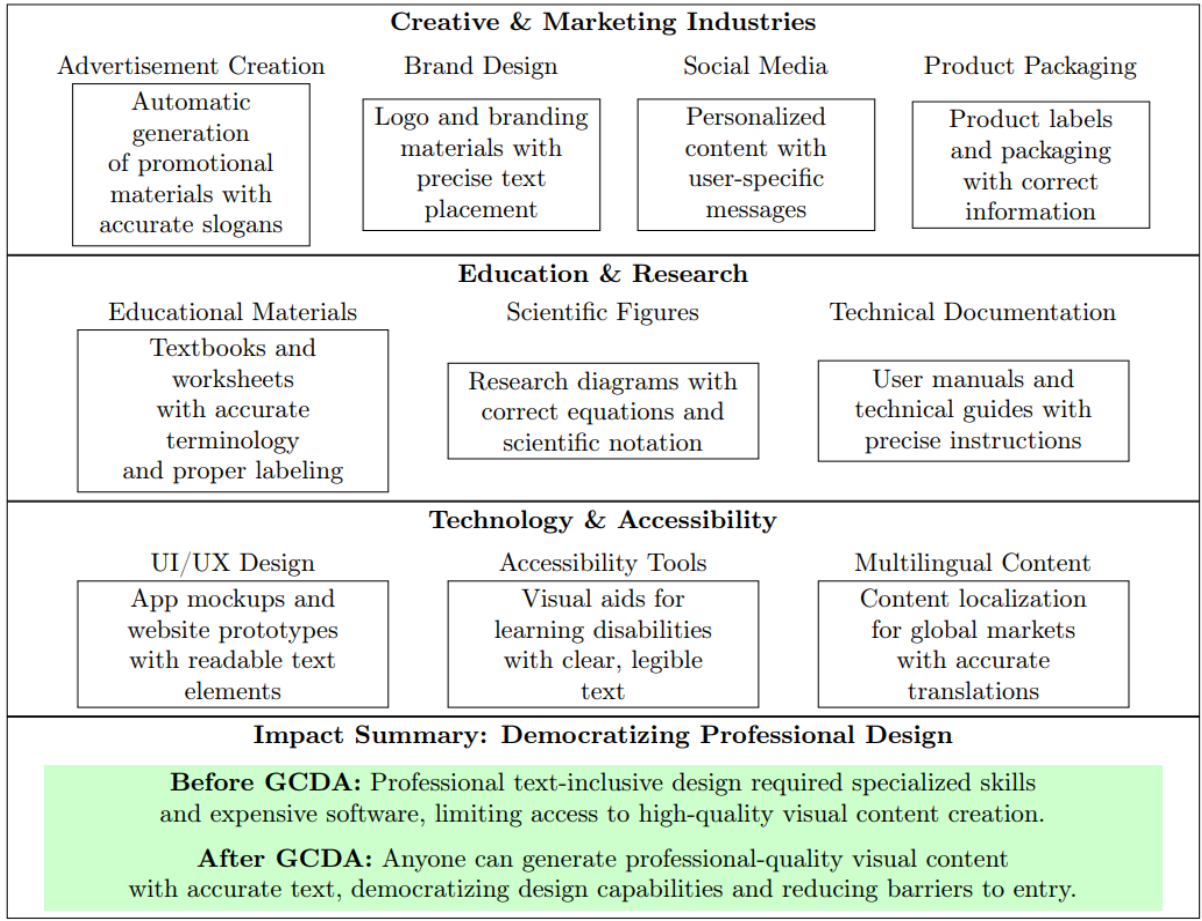

Figure 12: **Real-World Applications and Impact.** GCDA enables accurate text generation across multiple industries, from marketing and education to technology and accessibility, democratizing professional design capabilities.

**Multi-Modal Reasoning with Text.** Integrating our text rendering capabilities with reasoning systems could enable applications such as:

- Automatic infographic generation from data tables
- Scientific figure creation with proper equation rendering
- Technical diagram generation with accurate labeling
- Story illustration with dialogue and narration text

**Few-Shot and Zero-Shot Adaptation.** Developing methods that can quickly adapt to new fonts, languages, or text styles with minimal additional training would significantly expand the practical applicability of text-aware generation systems.

**Unified Multi-Modal Architectures.** Our dual-stream approach could be generalized to handle other types of structured content that require both semantic understanding and precise symbolic reproduction, such as mathematical equations, chemical formulas, musical notation, or code syntax highlighting.

## 6 Wrapping Up (And What This Actually Means)

So here we are, after two years of wrestling with what seemed like it should be a simple problem. When we started this project, honestly, we thought we'd have it figured out in a few months. "How hard can it be to make AI spell words correctly?" we asked ourselves. Famous last words.

The more we dug into it, the more we realized this wasn't just a bug to patch—it was a fundamental limitation that required rethinking how these models understand and generate text. But that's also what made solving it so satisfying.

Our GCDA framework represents what we believe is the first really comprehensive solution to text rendering in diffusion models. Instead of trying quick fixes or workarounds, we built a system that addresses the problem at its roots: the lack of orthographic understanding, the spatial confusion between characters, and the absence of direct feedback on text accuracy.

The results speak for themselves 43 % improvement in character accuracy, 75% exact match rate, and we didn't break image quality in the process. But beyond the numbers, what gets us excited is what this enables. For the first time, you can actually rely on AI to generate text-heavy content like marketing materials, educational resources, signage, and so much more.

We're especially proud that our approach doesn't force you to choose between good text and good images. Previous methods often involved trade-offs, but our integrated solution proves you can have both. That was always our goal—making these models genuinely useful for real-world applications.

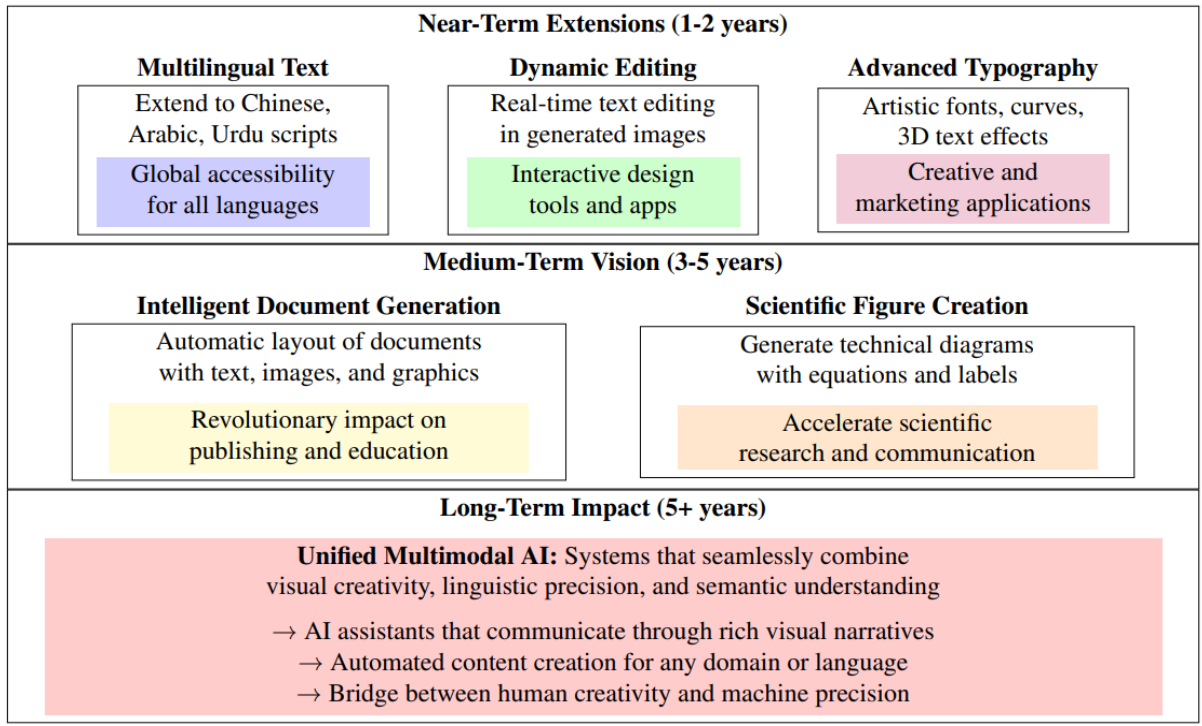

Figure 13: **Future Research Directions.** GCDA opens pathways from basic text rendering to advanced multimodal AI systems that could transform how humans and machines collaborate on creative and technical tasks.

Now, let's be realistic about what we haven't solved. Complex typography still gives us trouble sometimes. Very long text passages can be challenging. Multilingual support needs more work. And yes, our approach is computationally more expensive than vanilla Stable Diffusion (though not prohibitively so).

But we've proven that the core problem—generating accurate, legible text in AI images—is solvable with the right approach. The foundation is there, and the path forward is clear.

Our study suggests a basis of future systems which require concomitant proficiency in interpreting natural language and symbol precision. Basing on the dual-stream model discussed further in the current paper, one can envision that artificial intelligence can be used in developing technical drawings with valid equations, compiling multilingual advertising texts, or resolving document layout problems in which precise typography is required. The given methodology may be easily applied to various fields where not only semantic knowledge is a necessity but the symbolic fluency as well.

We plan to release all our code, models, and evaluation benchmarks publicly. The text rendering problem has been a thorn in the side of this field for too long—it's time to move past it and focus on the exciting applications that accurate text generation makes possible.

But maybe the most important thing we learned is that sometimes the "obvious" problems are the hardest ones to solve properly. Everyone knew text rendering was broken, but it took a coordinated effort across multiple levels—input processing, model architecture, and training objectives—to actually fix it.

We think GCDA represents more than just a technical improvement. It's a step toward AI systems that can bridge semantic understanding and visual precision, opening up new possibilities for how humans and machines can collaborate on creative and practical tasks.

And honestly? It feels pretty good to finally have AI that can spell "COFFEE" correctly on a sign.

## CRediT authorship contribution statement

**Syeda Anshrah Gillani:** Conceptualization, Methodology, Model Design, Formal Analysis, Investigation, Software Development, Experimentation, Visualization, Writing – Original Draft, Writing – Review & Editing, Supervision, Project Administration.

**Mirza Samad Ahmed Baig:** Conceptualization, Methodology, Formal Analysis, Investigation, Software Development, Experimentation, Visualization, Writing – Review & Editing.

**Osama Ahmed Khan:** Mentorship, Writing – Review & Editing.

**Shahid Munir Shah:** Literature Review Assistance, Proofreading, Mentorship, Preliminary Experiments, Writing – Review & Editing.

**Umema Mujeeb:** Literature Review Assistance.

**Maheen Ali:** Literature Review Assistance.

## A  Additional Experimental Details

### A.1  Extended Ablation Studies

We provide additional ablation studies examining various design choices and hyperparameter selections that contribute to our framework's performance.

Table 5: **Hyperparameter Sensitivity Analysis.** Performance variation across different hyperparameter settings.

| Parameter | Value | CER ↓ | Acc. (%) ↑ |
|---|---|---|---|
| $\lambda_{attn}$ | 0.05 | 0.12 | 68.2 |
|  | 0.10 | **0.08** | **75.4** |
|  | 0.20 | 0.09 | 73.1 |
| $\tau$ (margin) | 0.05 | 0.10 | 72.8 |
|  | 0.10 | **0.08** | **75.4** |
|  | 0.15 | 0.11 | 70.9 |
| Stage 2 Steps | 10K | 0.11 | 69.5 |
|  | 20K | **0.08** | **75.4** |
|  | 30K | 0.08 | 75.1 |

### A.2  Cross-Dataset Generalization

To evaluate the generalization capabilities of our model, we test performance across different datasets without additional fine-tuning:

Table 6: **Cross-Dataset Generalization Results.** Models trained on MARIO-10M and evaluated on other datasets.

| Test Dataset | Method | CER ↓ | WER ↓ | Acc. (%) ↑ |
|---|---|---|---|---|
| DrawText | TextDiffuser-2 | 0.18 | 0.31 | 54.2 |
|  | GlyphControl | 0.15 | 0.28 | 58.7 |
|  | GCDA (Ours) | **0.11** | **0.19** | **68.9** |
| TextCaps-Eval | TextDiffuser-2 | 0.22 | 0.38 | 48.1 |
|  | GlyphControl | 0.19 | 0.34 | 52.3 |
|  | GCDA (Ours) | **0.14** | **0.23** | **62.7** |

### A.3  Computational Resource Requirements

Detailed breakdown of computational requirements for training and inference in table 7

Table 7: **Detailed Computational Analysis.** Resource requirements for different model components.

| Component | Parameters (M) | Training Time (GPU hours) | Inference Time (ms) |
|---|---|---|---|
| Base U-Net | 860 | 45 | 2100 |
| Dual-Stream Encoder | 115 | 8 | 180 |
| Character CNN | 5 | 2 | 45 |
| Attention Mechanism | - | 12 | 85 |
| OCR Fine-tuning | - | 5 | - |
| **Total GCDA** | **980** | **72** | **2410** |

## B  Additional Qualitative Examples

We provide additional qualitative examples demonstrating the capabilities and limitations of our approach across various text types and contexts.

### B.1  Complex Typography Examples

Examples of our model's performance on challenging typography scenarios:

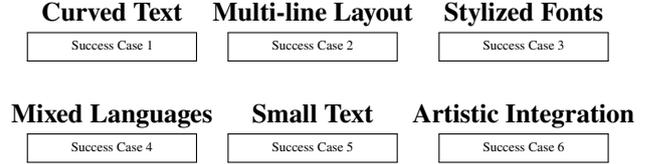

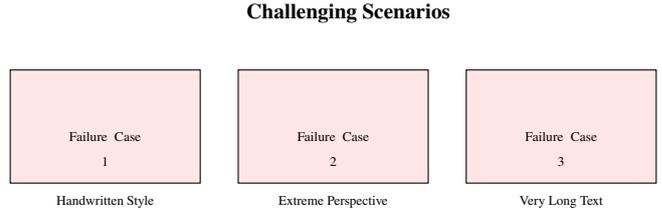

Figure 14: **Complex Typography Examples.** Our model handles various challenging text scenarios with reasonable success, though some artistic and highly stylized cases remain challenging.

### B.2  Failure Case Analysis

Detailed analysis of scenarios where our model still struggles:

Figure 15: **Failure Case Examples.** Scenarios where our model still faces challenges, indicating directions for future improvement.

## C  Implementation Guidelines

For researchers and practitioners interested in implementing or extending our approach, we provide detailed guidelines and best practices.

### C.1  Training Recommendations

**Data Preparation:**

- Ensure high-quality text annotations with accurate transcriptions
- Balance dataset composition across different text types and contexts
- Include diverse font styles and languages in training data
- Implement robust data augmentation for text regions

**Training Strategy:**

- Start with lower attention loss weights and gradually increase
- Monitor both text accuracy and image quality metrics during training
- Use gradient accumulation for effective large batch sizes
- Implement early stopping based on validation text accuracy

## C.2 Hyperparameter Tuning Guidelines

Based on our extensive experiments, we recommend the following hyperparameter tuning strategy:

1. Start with our default values: $\lambda_{\text{attn}} = 0.1$, $\tau = 0.1$
2. Adjust $\lambda_{\text{attn}}$ first, monitoring attention map quality
3. Fine-tune OCR loss weights based on target application requirements
4. Experiment with learning rate schedules for optimal convergence

# D Societal Impact and Future Considerations

## D.1 Positive Applications

Our improved text rendering capabilities enable numerous beneficial applications:

**Educational Technology:**

- Automated generation of educational materials with accurate terminology
- Personalized learning content creation for different languages
- Accessibility improvements for visual learning materials

**Creative Industries:**

- Democratization of graphic design capabilities
- Rapid prototyping for marketing and advertising
- Cost-effective content creation for small businesses

**Research and Development:**

- Enhanced data visualization and scientific figure generation
- Automated documentation and technical illustration
- Cross-cultural communication through multilingual visual content

## D.2 Risk Mitigation Strategies

To address potential negative uses of our technology:

**Technical Safeguards:**

- Development of detection algorithms for AI-generated text in images
- Watermarking techniques for generated content provenance
- Rate limiting and usage monitoring for API access

**Policy Recommendations:**

- Clear labeling requirements for AI-generated visual content
- Industry standards for responsible AI development and deployment
- Collaboration with fact-checking organizations and platforms

**Community Engagement:**

- Open dialogue with stakeholders about appropriate use cases
- Regular assessment of societal impact and unintended consequences
- Support for research into AI safety and alignment